%% file: linear.tex
\documentclass[10pt]{article}
\usepackage{iclr2027_conference,times}
\input{math_commands}

\usepackage{hyperref}       
\usepackage{url}            
\usepackage{booktabs}       
\usepackage{nicefrac}       
\usepackage{microtype}      
\usepackage{xcolor}         
\usepackage{graphicx}       
\usepackage{adjustbox}      
\usepackage{enumitem}       
\usepackage{listings}
\usepackage{subcaption}
\usepackage{wrapfig}
\usepackage{caption}
\usepackage{amssymb}
\usepackage{multirow}
\usepackage{amsthm}
\usepackage{longtable}
\usepackage{array}
\usepackage{arydshln}
\usepackage{tcolorbox}
\usepackage{mathtools}
\usepackage{float}
\usepackage{tikz}

\newtheorem{corollary}{Corollary}
\newtheorem{lemma}{Lemma}
\newtheorem{definition}{Definition}
\newtheorem{theorem}{Theorem}

\newsavebox{\leftcolbox}
\newcommand{\refAppendix}[1]{Appendix~\ref{#1}}

\title{High-probability guarantees for linear accessibility in feature superposition}

\author{Enrico Vompa (envomp@taltech.ee)\\
Applied Artificial Intelligence Group \\
Tallinn University of Technology, Estonia}

\iclrfinalcopy
\begin{document}

    \maketitle

    \vspace{-1.5em}
    \begin{abstract}
        \vspace{-0.4em}
        Neural networks can leverage feature superposition to encode more concepts than dimensions,
        but cross-feature interference constrains the linear accessibility of simultaneously active features.
        By framing linear accessibility as a compressed sensing problem,
        we derive high-probability recovery bounds for fixed supports under subgaussian noise,
        proving that $d = O_{\varepsilon}(k \log m)$ dimensions suffice for the per-input regime in which networks typically operate in,
        in contrast to the quadratic scaling required by uniform worst-case guarantees.
        We characterize the asymmetry between active and inactive interference and the trade-off between interference and observation-noise budgets.
        We then validate these bounds across system parameters through numerical experiments based on Gaussian-tail approximations.
        We also introduce IHT-SAE, which uses learned iterative refinement to improve feature recovery beyond the limits of linear accessibility.
        These results quantify the geometric constraints of the linear representation hypothesis,
        providing a framework for evaluating sparse autoencoders, compositional generalization, and neural interpretability.
        Code to reproduce this work is available at: \href{https://anonymous.4open.science/r/Linear-Availability-07CD}{https://anonymous.4open.science/r/Linear-Availability-07CD}
    \end{abstract}

    \vspace{-1.2em}

    \section{Introduction}\label{sec:introduction}

    Stochastic separation theorems show that in high dimensions,
    any point in a random set can be separated from the others by a hyperplane with high probability,
    even when the set grows exponentially with dimension~\citep{sidorov2020}.
    Though traditionally applied to single features, these concentration principles extend to simultaneously active features.
    Compressed sensing formalizes how a vast number of features ($m$)
    can be encoded into a smaller $d$-dimensional space when a sparse subset of size $k$ is active;
    mirroring feature superposition in neural networks~\citep{elhage2022superposition}.

    Even though LLM spaces are globally not uniform~\citep{ethayarajh2019contextual},
    concept subspaces are~\citep{cai2021isotropy},
    and can be effectively approximated by Gaussian distributions~\citep{zhao2023beyond}.
    It is this very local uniformity that gives rise to representational geometry (linear representation hypothesis)
    in vision encoders~\citep{radford2021learning} and LLMs~\citep{park2025the, park2024lrh},
    which encode features into composite structures (e.g., categorical, hierarchical, spatial, relational, etc).

    Within this geometry, co-occurring features such as ``Christmas'' and ``December'' can occupy similar directions,
    producing constructive interference that aids decoding~\citep{prieto2026from},
    while destructive interference can suppress active features~\citep{stevinson2025adversarial}.
    Yet true compositional reasoning may require recovering unusual combinations such as ``Christmas in July'' without familiar co-occurrences,
    making control of cross-feature interference central to reliable linear decoding.

    We apply the subgaussian Hoeffding inequality to derive high-probability bounds for linear decoding,
    showing that the sufficient dimension scales linearly with the number of active features,
    compared with quadratic scaling under uniform worst-case guarantees~\citep{garg2026featureslanguagemodelstore}.
    We further use Gaussian-tail approximations to estimate failure rates and assess these predictions numerically.

    Sparse autoencoders (SAEs) recover such features by reconstructing representations as sparse combinations of learned feature directions~\citep{huben2024sparse}.
    Standard single-step encoders, however, remain constrained by linear accessibility.
    We therefore introduce IHT-SAE, which refines a learned sparse estimate through residual-based updates with learned step sizes and TopK thresholding.
    We further find that decoder performance depends on the coefficient distribution:
    strongly decaying language-model activations favor greedy residual subtraction,
    while flatter vision-encoder activations narrow the gap between strategies.

    \newpage

    Our main contribution is a framework for understanding feature accessibility in superposition:
    \begin{itemize}[noitemsep, topsep=0pt, parsep=2pt, partopsep=0px, leftmargin=*]
        \item[-] We distinguish uniform worst-case from fixed-support high-probability linear accessibility guarantees,
        showing that in the latter, governed by cross-feature interference, a dimension $d = O_{\varepsilon}(k \log m)$ suffices for fixed error and failure tolerances
        when the noise satisfies $\sigma = O_{\varepsilon}(1 / \sqrt{\log m})$.
        \item[-] We place linear/nonlinear encoding/decoding within a shared representational regime, consistent with the linear representation hypothesis.
        Here, nonlinearities are used for setting/getting the feature coefficients to minimize or bypass the cross-feature interference.
        \item[-] We connect these guarantees to SAEs,
        introduce IHT-SAE to recover features beyond the limits of linear accessibility,
        while decoupling iteration count from the sparsity budget,
        and demonstrate that the choice of nonlinear decoder should match the model's coefficient distribution.
    \end{itemize}

    \section{Theory: compressed sensing framework}

    We model the problem in a compressed sensing framework, where a high-dimensional sparse vector is mapped to a lower-dimensional measurement space.

    \subsection{Linear versus nonlinear encoding}

    \begin{definition}[Linear encoding with independent noise]
        Let $x \in [-1, 1]^m$ be a $k$-sparse input signal,
        let $A = [a_1, \dots, a_m] \in \mathbb{R}^{d \times m}$ be a dictionary of unit $\ell_2$-norm feature embeddings ($\|a_i\|_2 = 1$),
        and let $z \in \mathbb{R}^d$ denote an independent observation noise vector.
        The linear encoding maps the input and noise to a compressed $d$-dimensional sketch $y$ via:
        $$y = Ax + z = \sum_{j \in S} x_j a_j + z$$
        where $S \subseteq [m]$ represents the active feature support of size $|S| \le k$.
        \label{def:linear_encoding}
    \end{definition}

    An alternative to this framework is encoding nonlinear representations.
    Linear representations possess a wider margin of error during decoding, operating as a probabilistic structure.
    Geometrically, decoding a linear sketch is akin to measuring the cosine similarity between nearly orthogonal feature directions;
    slight measurement errors yield proportionally small perturbations in the recovered state.
    In contrast, discontinuous nonlinear representations,
    such as RAM of storing the $k$ largest coefficients alongside their indices,
    can achieve a high compression of $d = O(k)$~\citep{ba2011lower}.
    However, slight errors during decoding of such structures (e.g., a corrupted index mapping) can completely break the structure,
    resulting in disparate representations rather than proportional deviations.
    Such a non-differentiable representation space renders gradient-based optimization intractable (but not impossible).
    We later show how nonlinearity can be employed for encoding linear representations while minimizing the margin of error when reading said features.

    \subsection{Nonlinear decoding}

    Classical compressed sensing demonstrates that a $k$-sparse vector in $\mathbb{R}^m$ can be embedded into approximately $d = O(k \log(m/k))$ measurements,
    assuming a known measurement matrix~\citep{candes2005decoding}.
    Here, recovery can be achieved in polynomial time using nonlinear decoding algorithms ($\ell_1$-minimization)~\citep{candes2006stable},
    which extends to noisy environments, provided the noise $z$ is bounded by $||z||_{\ell_2} \le \varepsilon$~\citep{candes2008ripimplications}.
    Assuming the measurement matrix satisfies the restricted isometry property with a constant $\delta_{2k} < \sqrt{2} - 1$,
    this recovery is guaranteed to be stable in both the $\ell_1/\ell_1$ and $\ell_2/\ell_1$ regimes~\citep{candes2008ripimplications},
    with a matching lower bound in these regimes~\citep{ba2011lower}.
    Here, these proofs hold uniformly for all combinatorially many $k$-sparse supports.

    \begin{figure}[H]
        \centering
        \includegraphics[width=\textwidth]{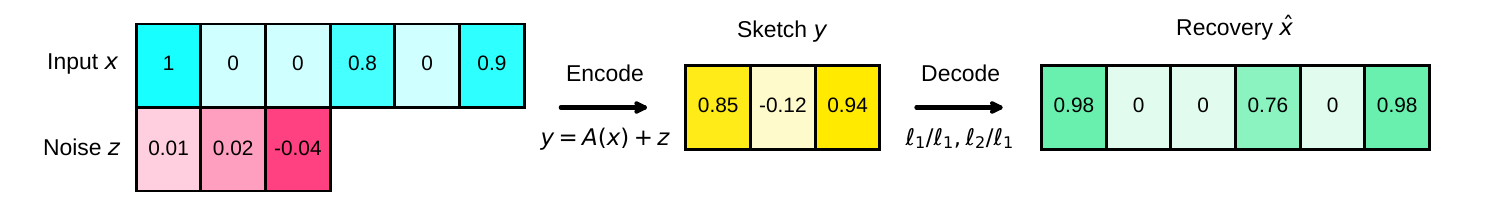}
        \caption{Compressed sensing encoding and decoding pipeline.}
        \label{fig:nonlinear_decoding}
    \end{figure}


    \subsection{Linear decoding}

    Linear decoding requires features to be linearly accessible, meaning they are recovered by a fixed linear decoder.
    As opposed to relatively clean recovery as in nonlinear decoding,
    linear decoding may produce cross-feature interference on both inactive and active coordinates~\citep{stevinson2025adversarial}.
    Depending on its sign, this interference can create spurious inactive estimates or distort active estimates.
    Decoding can therefore be framed as the existence of a threshold which separates active features from inactive ones.

    \begin{definition}[Linear decoding]
        Given a compressed sketch $y \in \mathbb{R}^d$ and a dictionary $A \in \mathbb{R}^{d \times m}$,
        linear decoding is defined by the single-step matched-filter projection:
        $$\widehat{x} = A^\top y$$
        such that the estimated coefficient for any feature $i \in [m]$ is recovered via the inner product $\widehat{x}_i = \langle a_i, y \rangle$.
    \end{definition}

\begin{figure}[H]
        \centering
        \includegraphics[width=\textwidth]{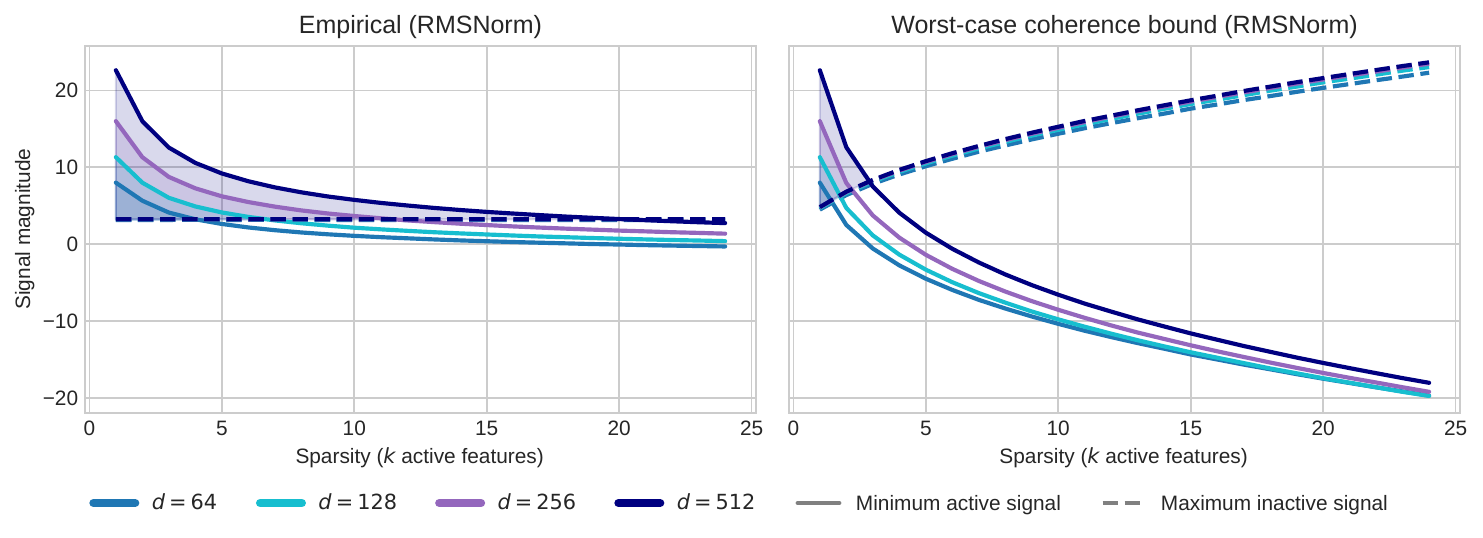}
        \caption{Signal magnitudes and recovery bounds in an RMSNorm vector space.}
        \label{fig:gap_rmsnorm}
    \end{figure}

    The geometric limits of linear decoding are invariant to the magnitude of the compressed state $y$.
    By the bilinearity of the inner product,
    scaling the compressed state by a factor $\alpha > 0$ scales the entire decoding projection equally $\langle a_i, \alpha y \rangle = \alpha \langle a_i, y \rangle$.
    Because the compressed state is a linear combination of active features ($y = \sum x_j a_j$),
    this scaling magnifies both the target signal and the interference by the same factor.
    Figure~\ref{fig:gap_rmsnorm} visualizes the recovery bounds in the empirical high-probability and worst-case scenarios,
    where a separating threshold could reside.

    \subsection{The worst-case scenario of linear decoding}
    Requiring the linear decoder to satisfy $\|A^\top Ax-x\|_\infty<\varepsilon$ uniformly
    for every $k$-sparse vector $x\in[-1,1]^m$ yields an upper bound $d=O_\varepsilon(k^2\log m)$;
    a nearly matching lower bound of $d=\Omega_\varepsilon\!\left(\frac{k^2}{\log k}\log\frac{m}{k}\right)$
    applies even when allowing any arbitrary linear decoder $B^\top$~\citep{garg2026featureslanguagemodelstore}.
    This $k^2$ scaling is a geometric limitation representing how effectively low-rank projections
    can suppress cross-feature interference uniformly over all combinatorially many $k$-sparse inputs.

    \subsection{The empirical high-probability linear decoding scenario}

    By analyzing the joint probability space of the dictionary, signal, and noise,
    we can bound the total cross-feature interference and noise.
    We partition the error tolerance $\varepsilon$,
    allocating a budget of $\alpha \in (0, 1)$ to noise and $(1-\alpha)$ to interference.
    A subgaussian tail bound controls both the interference $I_i$ and the projected noise.
    Because subgaussian tails decay exponentially fast,
    applying a union bound across all features guarantees with high probability that both components remain bounded within allocated budget.

    To bound the maximum interference, we partition the sum of individual tail probabilities into two:
    \begin{equation}
        \begin{aligned}
            &\mathbb{P}_{A,x}\left(\max_{i\in\mathcal I}|I_i|\ge(1-\alpha)\varepsilon\right)\\
            &\quad \le \underbrace{2(m-|S|)\exp\left(-\frac{d(1-\alpha)^2\varepsilon^2}{2|S|}\right)}_{\text{inactive}}
            +\underbrace{\mathbb{I}_{|S| \ge 2}\cdot 2|S|\exp\left(-\frac{d(1-\alpha)^2\varepsilon^2}{2(|S|-1)}\right)}_{\text{active}} \\
            &\quad \le2m\exp\left(-\frac{d(1-\alpha)^2\varepsilon^2}{2|S|}\right)\le\delta_{\text{int}}
        \end{aligned}
        \label{eq:joint_interference_terms}
    \end{equation}
    The inactive term bounds the interference on inactive coordinates,
    where the interference is generated by all $|S|$ active features,
    whose erroneous inclusion would produce false positives.
    The active term bounds the interference on active coordinates,
    where the interference is generated by the remaining $|S|-1$ active features,
    whose erroneous exclusion would produce false negatives.
    The union bound includes both terms,
    so the condition controls the interference failures relevant to both error types.

    \begin{lemma}[Interference constraint]
        Let $d, m, k \in \mathbb{N}_{>0}$ with $1 \le |S| \le k \le m$ for an arbitrarily chosen index set $S \subseteq [m]$.
        Suppose the dictionary columns $A = [a_1, \dots, a_m]$ are mean-zero and sampled independently and uniformly from the unit sphere $S^{d-1}$,
        and the active signal coefficients $(x_j)_{j \in S}$ are drawn from a joint distribution $P_x \in \mathcal{P}_S$
        satisfying $|x_j| \le 1$ independently of $A$ (with $x_j = 0$ for $j \notin S$).
        For any error threshold $\varepsilon > 0$, fraction $\alpha \in (0, 1)$, and failure probability $\delta_{\text{int}} \in (0, 1)$,
        if the dimension $d$ satisfies:
        $$d \ge \frac{2 |S|}{(1-\alpha)^2\varepsilon^2} \ln\left(\frac{2m}{\delta_{\text{int}}}\right)$$
        then the maximum cross-feature interference is bounded by $(1-\alpha)\varepsilon$ with probability at least $1 - \delta_{\text{int}}$.
        \label{lemma:interference_constraint}
    \end{lemma}
    For fixed $\alpha$ and $\delta_{\text{int}}$ independent of $m, k, \varepsilon$,
    substituting $|S| \le k$ yields the dimension scaling $d_{\text{req}} = O_{\varepsilon}(k \log m)$,
    matching the asymptotic scaling in hyperdimensional computing theory~\citep{thomas2021hypercomputing}.
    However, unlike prior work that absorbs coefficients into feature vectors (implicitly assuming unit scale),
    we separate feature directions from their coefficient distributions,
    where our guarantees hold for arbitrary bounded coefficient distributions.
    This separation also yields a variance proxy for interference (Equation~\ref{eq:variance_proxy});
    combined with the noise variance, this proxy explains the budget borrowing dynamic.
    It also allows us to distinguish active from inactive interference.

    \begin{lemma}[Noise constraint]
        Let the dictionary $A$ and signal $x$ satisfy the conditions of Lemma~\ref{lemma:interference_constraint}.
        Suppose the observation noise $z = (z_1, \dots, z_d)$ is drawn from a product distribution $P_z \in \mathcal{Z}_\sigma$ independent of $(A, x)$,
        with independent, mean-zero coordinates satisfying the subgaussian norm bound $\|z_n\|_{\psi_2} \le \sigma$ for $\sigma > 0$.
        For any fraction $\alpha \in (0, 1)$, error tolerance $\varepsilon > 0$, and noise failure budget $\delta_{\text{noise}} \in (0, 1)$,
        if the subgaussian scale satisfies:
        $$\sigma \le \frac{\sqrt{c}\alpha\varepsilon}{\sqrt{\ln(2m/\delta_{\text{noise}})}}$$
        then noise is bounded by $\alpha\varepsilon$ with probability at least $1 - \delta_{\text{noise}}$,
        where $c > 0$ is an absolute constant.
        \label{lemma:noise_constraint}
    \end{lemma}
    For fixed $\alpha$ and $\delta_{\text{noise}}$, this bounds the maximum certified noise scales as $\sigma_{\text{max}} = O_{\varepsilon}(1 / \sqrt{\log m})$.

    \begin{theorem}[Fixed-support decoding guarantee]
        Let $\varepsilon > 0$ and $\alpha \in (0, 1)$,
        and let the failure budget be partitioned into $\delta_{\text{int}}, \delta_{\text{noise}} \in (0, 1)$
        such that $\delta_{\text{int}} + \delta_{\text{noise}} = \Delta < 1$.
        If the dimension $d$ and noise scale $\sigma$ satisfy Lemma~\ref{lemma:interference_constraint} and Lemma~\ref{lemma:noise_constraint}, respectively,
        then:
        $$\forall\, S \subseteq [m]\ \text{with}\ 1 \le |S| \le k,\qquad
        \sup_{P_x \in \mathcal{P}_S} \sup_{P_z \in \mathcal{Z}_\sigma}
        \mathbb{P}_{A,x,z}\left(\|\widehat{x} - x\|_\infty \ge \varepsilon\right)
        \le \Delta$$
        \label{theorem:global_joint_guarantee}
    \end{theorem}
    Meaning, for any arbitrary fixed support $S$ across all valid distributions,
    the maximum estimation error exceeds $\varepsilon$ with a joint failure probability of at most $\Delta$.

    \begin{corollary}[High-probability support recovery]
        \label{cor:support_recovery}
        Given the $\ell_\infty$ error bound $\|\widehat{x} - x\|_\infty \le \varepsilon$ from Theorem~\ref{theorem:global_joint_guarantee}
        and a known minimum active coefficient $x_{\min} \le \min_{j \in S} |x_j|$,
        support recovery via the thresholding rule $\widehat{S} = \{i : |\widehat{x}_i| > \tau\}$ is guaranteed
        if there exists a threshold $\tau$ satisfying:
        $$\varepsilon_{\text{inactive}} < x_{\min} - \varepsilon_{\text{active}}\implies
        \underbrace{\varepsilon_{\text{inactive}}}_{\max_{i \notin S} |\widehat{x}_i|}
        < \tau < \underbrace{x_{\min} - \varepsilon_{\text{active}}}_{\min_{j \in S} |\widehat{x}_j|}$$
        \label{eq:exact_support_separate}
    \end{corollary}

    Full proof is in~\refAppendix{sec:appendix_average_case_proof}.

    \section{Numerical tests: evaluating theory}
    \label{sec:empirical_validation}

    In our numerical tests (unless specified otherwise),
    all dictionary columns are drawn from Gaussian distribution and L2-normalized to unit length.
    For each sample, $k$ active features are selected uniformly at random with coefficients set to $1$,
    yielding noiseless observations.

    As the Hoeffding inequality provides a conservative bound for any dictionary uniformly distributed on the unit sphere $S^{d-1}$,
    we can obtain a more precise estimate for the failure probability $\delta_{\text{gauss}}$
    by modeling the interference terms independently (Equation~\ref{eq:joint_interference_terms})
    and computing their Gaussian approximation (\refAppendix{subsec:theoretical_delta}).
    Because this formulation bounds the tail-matching factor from below by a constant ($0.5$),
    the Gaussian approximation occasionally falls below the Hoeffding bound but never exceeds it.
    This approximation effectively captures the scaling dynamics of the system parameters.
    In configurations where the number of features is small relative to the dimension size, it yields a high probability of successful decoding;
    and instead when dimension size is relatively smaller, the probability decreases accordingly.

    With minimum active coefficient $x_{\min} = 1$, the empirical maximum deviations are computed as
    $\max_{i \in S} |1.0 - \widehat{x}_i|$ and $\max_{i \notin S} |\widehat{x}_i|$ for active and inactive terms,
    respectively, from which we derive the empirical failure rate $\delta_{\text{emp}}$.
    As $\delta$ approaches zero, the term $\ln(2m/\delta)$ grows indefinitely.
    To prevent finite-sample Monte Carlo noise from distorting the failure-rate ratio $\delta_{\text{gauss}} / \delta_{\text{emp}}$,
    we restrict our evaluation to regimes with at least $100$ empirical failures.
    Because the empirical failure count $N_{\text{fail}}$ for rare events approximates a Poisson distribution,
    its relative statistical error (coefficient of variation) is defined
    by the ratio of the standard deviation to the mean, $\sigma/\mu \approx \sqrt{N_{\text{fail}}}/N_{\text{fail}} = 1/\sqrt{N_{\text{fail}}}$~\citep{rubinstein2016montecarlo}.
    By requiring $N_{\text{fail}} \ge 100$, the estimated coefficient of variation of the empirical denominator is at most $1/\sqrt{100} = 10\%$,
    reducing the influence of sampling noise on the failure-rate ratio.
    Meaning, $40,000$ trials enables evaluating high-probability decoding regimes
    down to a minimum failure rate of $0.25\%$ (a $99.75\%$ success rate).

\begin{figure}[H]
        \centering
        \includegraphics[width=\textwidth]{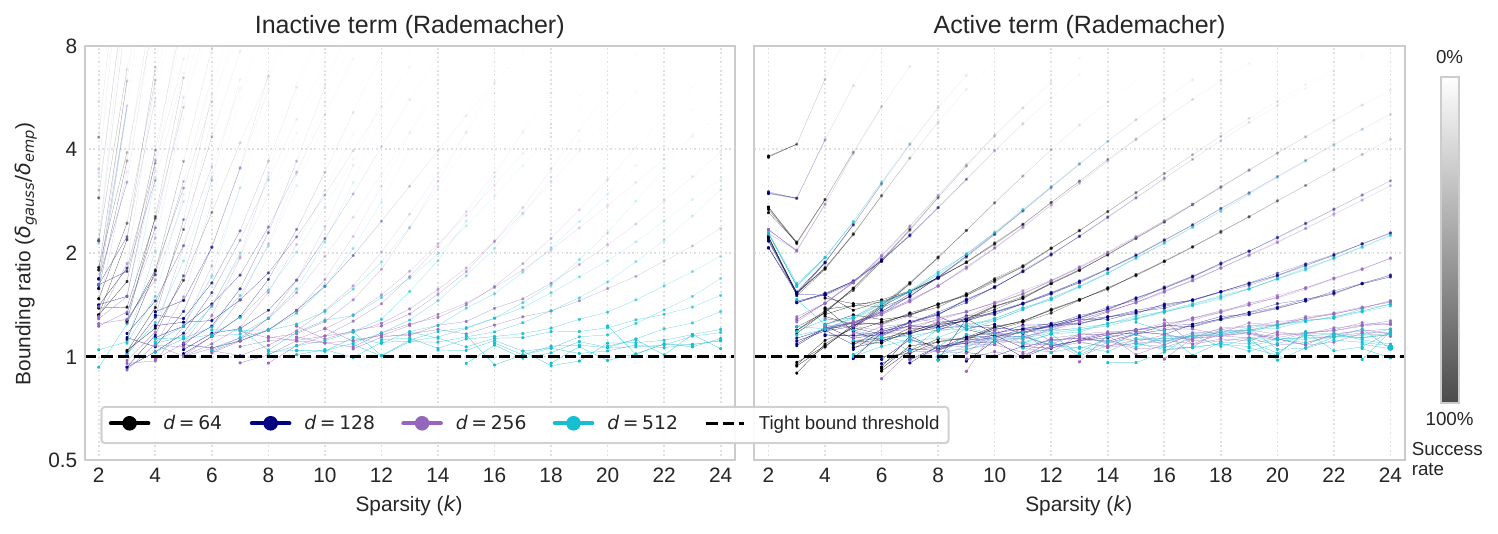}
        \vspace{-20pt}
        \caption{Tightness of Gaussian approximation for high-probability decoding on Rademacher distribution.
            $40,000$ trials were run for every combination of sparsity $k \in [2, 24]$,
            dictionary size $m \in \{512, 1024, 2048, 4096\}$ and dimension $d \in \{64, 128, 256, 512\}$,
            evaluated against various error thresholds $\varepsilon \in \{0.1, 0.2, \dots, 0.9\}$.
            When $k=2$ and active coefficients have equal magnitude,
            a projection $\langle a_i, a_j \rangle$ equally contributes to the failures of target $i$ and $j$,
            so the union bound counts the same underlying event twice (negligible for larger $k$).}
        \label{fig:rademacher_bound}
    \end{figure}

    As recovery success approaches 100\%, the $\delta_{\text{emp}}$ converges to $\delta_{\text{gauss}}$ (Figure~\ref{fig:rademacher_bound}),
    demonstrating that the Gaussian approximation closely estimates the required dimension for high-probability decoding,
    providing empirical support that Hoeffding inequality governs the guarantee for high-probability (low failure-probability) linear accessibility.
    Similar results for Gaussian and Laplace\footnote{Though Laplace distribution is subexponential,
        because we project the dictionary columns onto a unit sphere,
        it becomes bounded and subsequently subgaussian.
        These normalized columns are not uniform on the sphere, however.
        Nevertheless, under suitable central limit theorem conditions, centered,
        variance-normalized interference asymptotically approaches a normal distribution,
        motivating the Gaussian approximation.}
    distributions are shown in~\refAppendix{sec:appendix_gaussian_approximation_on_other_distributions}.

    \subsection{Support recovery}

    \begin{minipage}[t]{0.56\textwidth}
        Figures~\ref{fig:success_rate_spaces} and ~\ref{fig:success_rate_noise} evaluate success
        by the existence of a valid separating threshold (Corollary~\ref{cor:support_recovery}).

        \vspace{0.5em}
        In general vector space, where active features are simply summed together,
        the magnitude of the superposed state grows unbounded as the number of simultaneous features ($k$) increases.
        Subsequently, applying RMSNorm~\citep{zhang2019rmsnorm} constrains the superposed state vector to a fixed magnitude of $\sqrt{d}$.
        Despite this difference in magnitudes, relative signal-to-interference ratios are identical
        as normalization scales the target signal and the background geometric interference by the same factor.
    \end{minipage}
    \hfill
    \begin{minipage}[t]{0.42\textwidth}
        \vspace{-40pt}
        \centering
        \includegraphics[width=\linewidth]{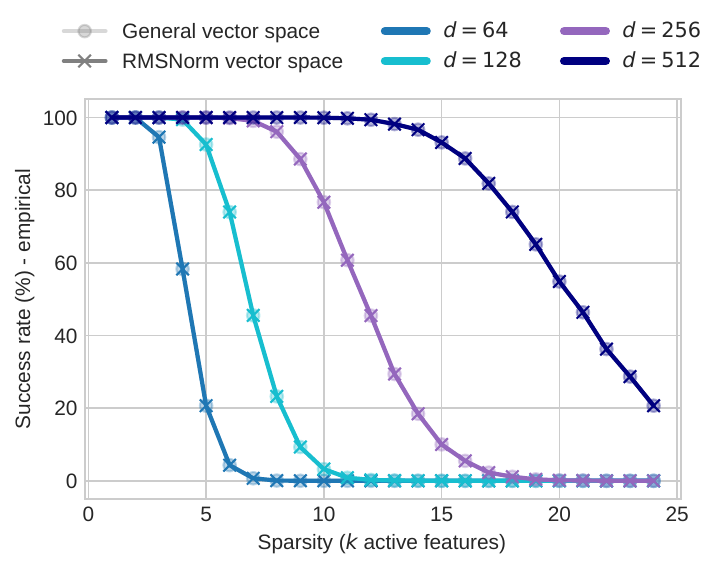}
        \vspace{-20pt}
        \captionof{figure}{Recovery of general and RMSNorm vector spaces across dimensions and sparsities.}
        \label{fig:success_rate_spaces}
    \end{minipage}

    \begin{minipage}[t]{0.56\textwidth}
        Our proofs partition the error tolerance $\varepsilon$ into noise budget ($\alpha\varepsilon$) and interference budget ($(1-\alpha)\varepsilon$).

        \vspace{0.5em}
        In practice, success simply requires that the sum of interference and noise remains below the tolerance $\varepsilon$.
        Because we model interference and additive noise as independent variables,
        Bienaymé's identity dictates that the variance of their sum equals the sum of their individual variances
        (we derive the variance proxy for interference in Equation~\ref{eq:variance_proxy}).
        Meaning, the system exhibits budget borrowing,
        where high interference can consume the total tolerance $\varepsilon$,
        leaving no room for noise, and vice versa.
    \end{minipage}
    \hfill
    \begin{minipage}[t]{0.42\textwidth}
        \vspace{-15pt}
        \centering
        \includegraphics[width=\linewidth]{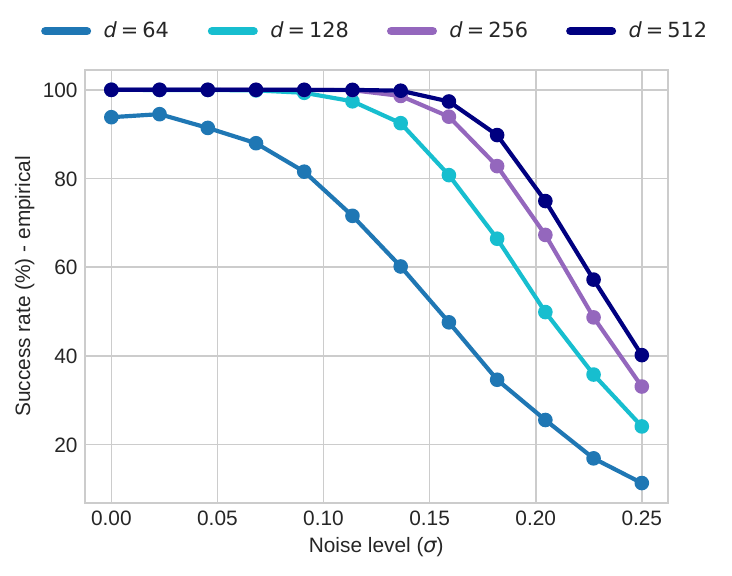}
        \vspace{-20pt}
        \captionof{figure}{Recovery in noisy environment.}
        \label{fig:success_rate_noise}
    \end{minipage}

%


    \vspace{-10pt}

    \subsection{Nonlinear decoding algorithms}

    When dictionary is known, sparse inference can be performed using methods such as OMP (orthogonal matching pursuit),
    iterative hard thresholding, or LASSO (least absolute shrinkage and selection operator).
    OMP greedily orthogonalizes the residual of linearly most accessible feature~\citep{pati1993omp}.
    Iterative hard thresholding instead uses residual correlations to correct all feature activation estimates,
    then retains the $k$ largest in magnitude~\citep{blumensath2009hard}.
    And LASSO estimates sparse coefficients by minimizing reconstruction error with an $\ell_1$ penalty~\citep{robert1996lasso}.
    Meaning, it operates in a nonlinear decoding regime where the theoretical bounds of $\ell_1/\ell_1$ apply.
    This convex problem can be solved using ISTA (iterative shrinkage-thresholding algorithm)~\citep{daubechies2004ista},
    which alternates residual-based coefficient updates with soft thresholding, or its accelerated variant, FISTA~\citep{beck2009fista}.

    In unsupervised dictionary learning,
    the optimization must simultaneously discover feature directions and identify the unknown $k$-sparse activation support,
    where coefficients can be estimated using different sparse-inference methods, such as OMP, FISTA, or ISTA (\refAppendix{sec:dictionary_learning}).
    Building on iterative soft thresholding, LISTA (learned iterative shrinkage-thresholding algorithm) unfolds ISTA
    into a fixed-depth network and learns its parameters through backpropagation~\citep{gregor2010lista}.

    \subsection{Feature recovery using SAEs}
    \label{subsec:feature_recovery_using_saes}

    A TopK SAE~\citep{gao2024topk} encoder resembles a learned hard-thresholding step,
    whereas a ReLU SAE encoder resembles LISTA without iterative refinement.
    Both amortize the cost of sparse inference via a learned encoder that predicts sparse codes in a single forward pass,
    jointly optimizing the encoder and decoder dictionary through gradient descent.
    As monotonic activation functions (e.g., ReLU) do not increase the capacity of linear accessibility~\citep{garg2026featureslanguagemodelstore},
    these standard SAEs operate in a linear decoding regime.
    This could explain why SAEs fail at compositional generalization within a compressed sensing framework when trained on LLM activations~\citep{pacela2026stop}.
    Because Transformers can provably implement LISTA-type iterative nonlinear decoding algorithms~\citep{liu2025on},
    they have the capacity to operate in a nonlinear decoding regime.
    Meaning, LLMs can represent features in denser superposition that remain inaccessible to the linear decoding regime of SAEs.

    To elevate SAEs into a nonlinear decoding regime,
    MP-SAE incorporates matching pursuit into an SAE by learning a dictionary
    through an unrolled matching-pursuit encoder that performs sequential residual updates~\citep{costa2025from}.
    To further improve robustness to unknown sparsity,
    building on iterative hard thresholding, we augment a TopK SAE with learned iterative refinement (as in LISTA),
    and tie the update matrices to the decoder dictionary in order to align refinement with the SAE’s reconstruction objective.

    \begin{figure}[H]
        \centering
        \includegraphics[width=\linewidth]{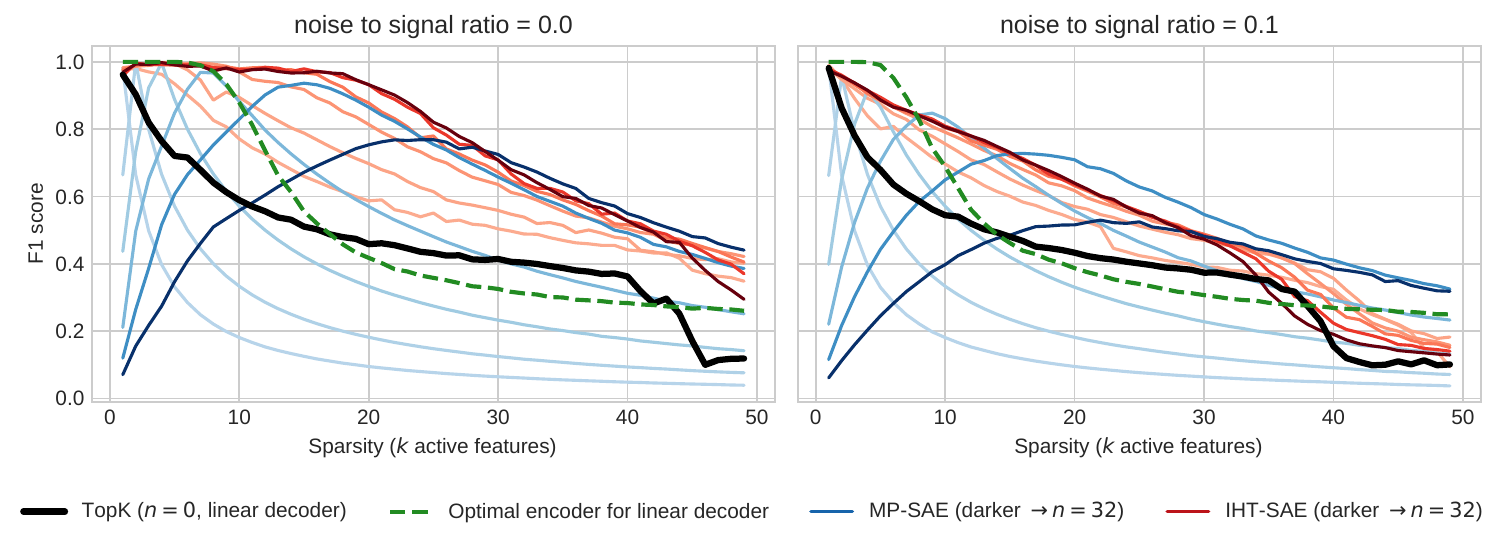}
        \captionof{figure}{Support recovery using TopK, IHT-SAE and MP-SAE in RMSNorm vector space ($m=1024$, $d=128$).
        Curves show F1 scores versus active sparsity $k$ at noise-to-signal power ratios of $0$ and $0.1$.
        To evaluate support recovery up to permutation and sign ambiguities,
            predicted feature activations are greedily aligned with the ground truth by mapping each true dictionary direction
            to the learned latent direction that maximizes their absolute cosine similarity.
            Reliable support recovery requires active coefficients to be sufficiently large to remain distinguishable from cross-feature interference and noise.
            We therefore impose a minimum active coefficient of $0.1$ before normalization.
            Darker shades indicate more iterations ($n \in \{1,2,4,8,16,32\}$); black denotes single-step TopK ($n=0$).
            The dashed green line denotes the optimal encoder baseline.
            Similar results for general vector space in~\ref{subsec:appendix_sae_general_vector_space}.
            Detailed methodology in Section~\ref{sec:methodology_sae}.}
        \vspace{-10pt}
        \label{fig:sae_comparison}
    \end{figure}

    Our IHT-SAE (algorithm in~\refAppendix{sec:appendix_IHT_sae}) refines a learned sparse initialization through residual-based updates with learned step sizes.
    Each update adjusts coefficients jointly, then retains at most top-$k$ values, followed by ReLU\@.
    MP-SAE instead starts from zero, greedily selecting a feature from the residual and subtracting its contribution.
    Thus, $n=k$ is an ideal iteration budget for MP-SAE if each iteration selects one new, correct feature.
    But choosing that $n$ comes with the caveat of needing to know the active sparsity.
    IHT-SAE decouples iteration count from the sparsity budget by retaining at most $k$ features after each update.

    We also compute an optimal coefficient baseline that maximizes the separation margin between active and inactive features
    using gradient descent (methodology in \refAppendix{subsec:optimal_encoder_baseline}),
    outperforming a regular linear decoder.
    For context, every iteration in isolation within these SAEs is governed by linear accessibility,
    so future work could incorporate better encoding to preemptively account for interference and boost recovery performance even further.
    This also extends to foundational models,
    where one could maximize linear accessibility while minimizing false positives and negatives.

    \begin{figure}[H]
        \centering
        \includegraphics[width=\textwidth]{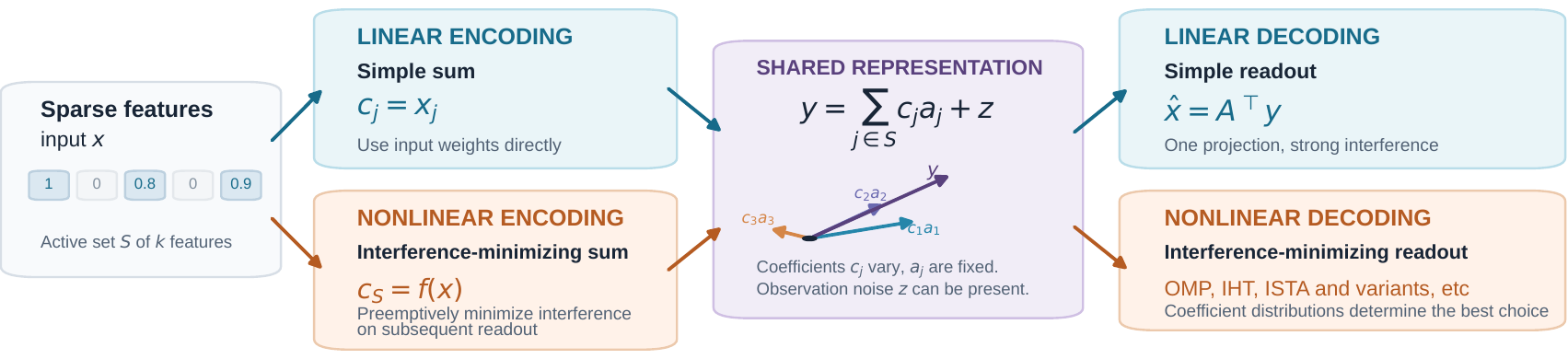}
        \caption{Overview of the compressed sensing framework of linear/nonlinear encoding/decoding.}
        \label{fig:framework}
    \end{figure}

    \newpage

    \subsubsection{Evaluation on foundation models}
    \label{subsubsec:empirical_evaluation_foundation_models}

    We evaluated both IHT-SAE and MP-SAE on activations extracted from foundational models.
    We probed layer index 18 of Qwen3-4B~\citep{bai2023qwentechnicalreport} using the TinyStories dataset~\citep{eldan2023tiny}
    and layer index 5 of the CLIP ViT-Base~\citep{radford2021learning} using the Oxford-IIIT Pet dataset~\citep{parkhi12a}.
    We report the reconstruction variance explained ($R^2$,~\citet{bricken2023monosemanticity}) and downstream performance preservation,
    measured by the degradation in cross-entropy (CE) loss when replacing original model activations with the SAE reconstructions.
    The training methodology mirrors the synthetic setup (Section~\ref{sec:methodology_sae}),
    where models were trained for $10,000$ steps with a batch size of $1024$ using the Adam optimizer,
    incorporating $\ell_2$-normalization of the decoder dictionary after every update.
    For these experiments, we fixed the sparsity budget to $k=32$ and set the SAE dictionary size to four times the hidden dimension of the respective foundation model.
    We evaluated single-step TopK ($n=0$), IHT-SAE ($n \in \{4, 8, 16\}$), and MP-SAE ($n=32$).
    Here, 16 iteration IHT-SAE and 32 iteration MP-SAE take equal amount of FLOPs per forward pass,
    where each IHT iteration takes roughly the amount of FLOPs of baseline TopK forward pass.

    \begin{table}[H]
        \centering
        \caption{Downstream performance and reconstruction variance explained of SAE architectures on foundation models.
        The baseline row shows the original cross-entropy (CE) loss.
        SAE rows report mean $\pm$ sample standard deviation across three seeds for $R^2$ reconstruction
        and the increase in CE loss ($\Delta$) when substituting original activations with SAE reconstructions.}
        \resizebox{\textwidth}{!}{
            \begin{tabular}{lcccc}
                \toprule
                & \multicolumn{2}{c}{CLIP ViT} & \multicolumn{2}{c}{Qwen3 4B GPT} \\
                \cmidrule(lr){2-3} \cmidrule(lr){4-5}
                Encoder             & $R^2$                        & $\Delta$ CE loss              & $R^2$                        & $\Delta$ CE loss              \\
                \midrule
                Baseline            & --                           & 1.6961                        & --                           & 1.9833                        \\
                \midrule
                TopK (0 iterations) & $0.7000 \pm 0.0003$          & $+0.0558 \pm 0.0036$          & $0.9989 \pm 0.0000$ & $+0.0996 \pm 0.0022$ \\
                IHT (4 iterations)  & $0.7509 \pm 0.0008$          & $+0.0286 \pm 0.0063$          & $0.9989 \pm 0.0000$          & $+0.0976 \pm 0.0026$          \\
                IHT (8 iterations)  & $0.7576 \pm 0.0010$          & $+0.0226 \pm 0.0010$          & $0.9989 \pm 0.0000$          & $+0.1252 \pm 0.0075$          \\
                IHT (16 iterations) & $0.7584 \pm 0.0024$          & $+0.0255 \pm 0.0022$          & $0.9988 \pm 0.0000$ & $+0.1318 \pm 0.0102$ \\
                MP (32 iterations)  & $\mathbf{0.7623 \pm 0.0001}$ & $\mathbf{+0.0201 \pm 0.0023}$ & $\mathbf{0.9990 \pm 0.0000}$ & $\mathbf{+0.0414 \pm 0.0008}$ \\
                \bottomrule
            \end{tabular}
            \label{tab:sae_reconstruction}}
    \end{table}

    Different nonlinear decoding algorithms handle coefficient distributions differently,
    meaning, the best algorithm depends on the coefficient distribution of the model being analyzed.
    OMP has been shown to perform well in $\alpha$-strongly-decaying coefficient distributions~\citep{wen2020coefficient},
    where the non-zero active coefficients,
    when sorted in descending order of magnitude such that $|x_{(1)}| \ge |x_{(2)}| \ge \dots \ge |x_{(k)}| > 0$,
    satisfy the condition:
    $|x_{(i+1)}| \le \alpha |x_{(i)}| \quad \text{for all } i \in \{1, \dots, k-1\}$.

\begin{figure}[H]
        \centering
        \begin{subfigure}{0.48\textwidth}
            \centering
            \includegraphics[width=\textwidth]{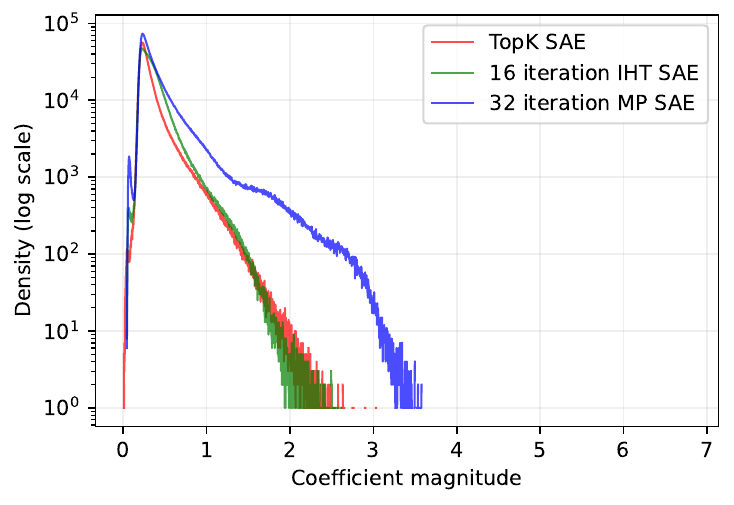}
            \caption{CLIP vision model}
            \label{fig:sae_clip}
        \end{subfigure}\hfill
        \begin{subfigure}{0.48\textwidth}
            \centering
            \includegraphics[width=\textwidth]{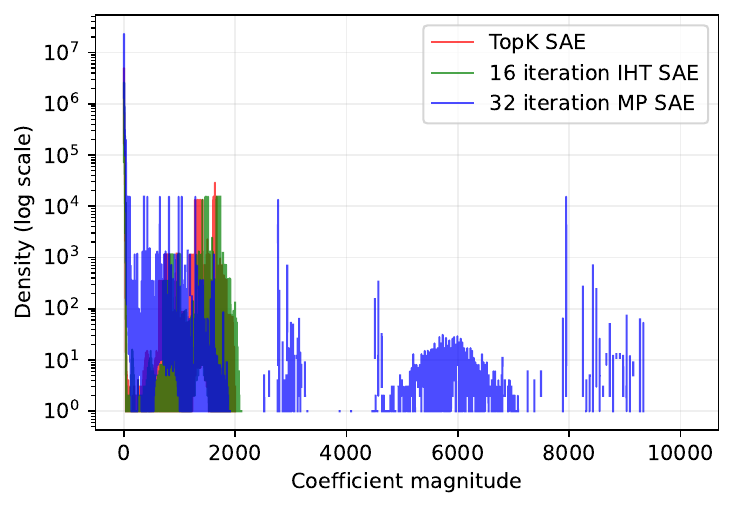}
            \caption{GPT language model}
            \label{fig:sae_gpt}
        \end{subfigure}
        \caption{Unfilled histogram of coefficient distributions on a logarithmic y-axis.
        MP SAE has an inductive bias towards an $\alpha$-strongly-decaying coefficient distribution,
            highlighting its capacity to isolate extreme outlier activations in language models.}
        \label{fig:sae_distributions}
    \end{figure}

    This allows it to greedily subtract massive outliers without them distorting the recovery of smaller features,
    which explains the better downstream reconstruction loss of MP-based SAEs over IHT in LLMs (Figure~\ref{fig:sae_distributions}).
    As such, we provide the theoretical and empirical pieces
    for constructing SAEs whose inductive bias matches that of the model being analyzed.

    \section{Discussion}\label{sec:discussions_limitation_future_work}

    It is important to distinguish a concept manifold from features.
    While a conceptual manifold, all things that could be (e.g., days of the week) might not be one-dimensionally linear,
    its materialized instances (e.g., Monday) exists as a linearly separable point along that structure~\citep{engels2025not}.
    While it has been shown to extend to higher dimensional manifold structures (numbers on a helix,~\citet{kantamneni2025helix}),
    every instance (e.g, Friday the second) is still a composite feature of sparse activations.

    \subsection{Unknown nonlinear transform}

    A parallel line of work, known as World Models, has emerged
    which force nonlinear observations into a linear transform of the world's latent structure~\citep{klindt2026does}.
    This is best understood through structural features,
    which are compositional properties that require nonlinearity to be recovered~\citep{vompa2026scaling}.
    Rather than assuming representations are inherently nonlinear,
    we hypothesize it is more akin to an unsolved problem where the base features are represented linearly,
    while higher-order properties are derived via iterative bottom-up perception (e.g., nonlinear encoding) or top-down reasoning (e.g., nonlinear decoding)~\citep{vompa2026beyond}.
    Here, new information is added linearly to the residual stream,
    while successive layers can transform the representational basis and refine feature estimates.
    This enables iterative nonlinear encoding/decoding,
    providing a mechanism through which greater Transformer depth can improve compositional generalization~\citep{vompa2026scaling}.

%

    \subsection{Limitations}

    The connection between linear accessibility and LLM mechanics remains speculative,
    relying on external empirical evidence.
    Under the conditions of the Lindeberg central limit theorem~\citep{lindeberg1922},
    centered, variance-normalized interference asymptotically converges to a normal distribution as $|S|$ grows;
    coupled with normalization (typically RMSNorm), their linear projections become bounded and subsequently subgaussian.
    This does not guarantee the uniform feature directions assumed by our theorem;
    nevertheless, we show that Gaussian tails are bounded by the Hoeffding bound theoretically and through numerical experiments.
    We describe the conditions where linear accessibility is possible,
    and subsequently the sufficient conditions for SAEs and more generally LRH,
    the extent LLMs utilize this remains unknown.
    This work does not claim a new state-of-the-art architecture,
    but rather clarifies the conditions governing a decoder choice.

    \section{Conclusion}\label{sec:conclusion}

    We characterize the high-probability guarantees of linear accessibility in feature superposition,
    showing that a sufficient dimension for fixed-support decoding is $d = O(k\log m)$ dimensions.
    Within this framework, we demonstrate that linear/nonlinear encoding/decoding can share a linear representational regime (assumed by LRH) of a simple weighted sum over active features,
    where nonlinearities can be used for setting/getting the coefficients to bypass cross-feature interference.
    These results clarify the geometric limits of the linear representation hypothesis and suggest how learned superposition can support compositional reasoning.

%

    \newpage

    \bibliographystyle{iclr2027_conference}
    \bibliography{references}

    \newpage

    \appendix

    \setcounter{table}{0}
    \setcounter{figure}{0}
    \renewcommand{\thetable}{A.\arabic{table}}
    \renewcommand{\thefigure}{A.\arabic{figure}}

    \newpage

    \input{linear_sufficient_average_case}

    \section{Dictionary recovery}\label{sec:dictionary_learning}

    To measure dictionary recovery across varying dimensions ($d \in \{64, 128, 256\}$) and sparsities ($k \in [1, 50)$) for a dictionary of size $m=1024$,
    we employ stochastic alternating minimization for up to 100 iterations using batches of $N = 50,000$ samples.
    The samples are generated via $Y = D_{\text{true}}X$ (general vector space),
    where $k$ active coefficients per sample are drawn from a normal distribution $\mathcal{N}(0,1)$ with uniformly random support.

    Sparse codes are estimated using OMP (constrained to $k$ non-zeros), FISTA (50 iterations with Nesterov acceleration and spectral norm-derived step sizes),
    and ISTA (unrolled for 10 iterations using weight matrices derived from the current dictionary),
    with both FISTA and ISTA optimizing the LASSO objective ($\lambda=0.5$).

    The dictionary is then updated from accumulated sufficient statistics $XX^T$ and $YX^T$ using ridge-regularized MOD with a penalty of $10^{-4}$,
    followed by $\ell_2$ column normalization~\citep{engan1999mod}.
    Degenerate atoms ($||d_i||_2 < 10^{-8}$) are dynamically reinitialized from random batch samples,
    and training halts early if the maximum column shift falls below $10^{-5}$.
    Recovery is evaluated up to sign and permutation ambiguities by applying the Hungarian matching algorithm to the cost matrix $-|D_{\text{true}}^T D_{\text{pred}}|$.

    \begin{figure}[H]
        \centering
        \includegraphics[width=\linewidth]{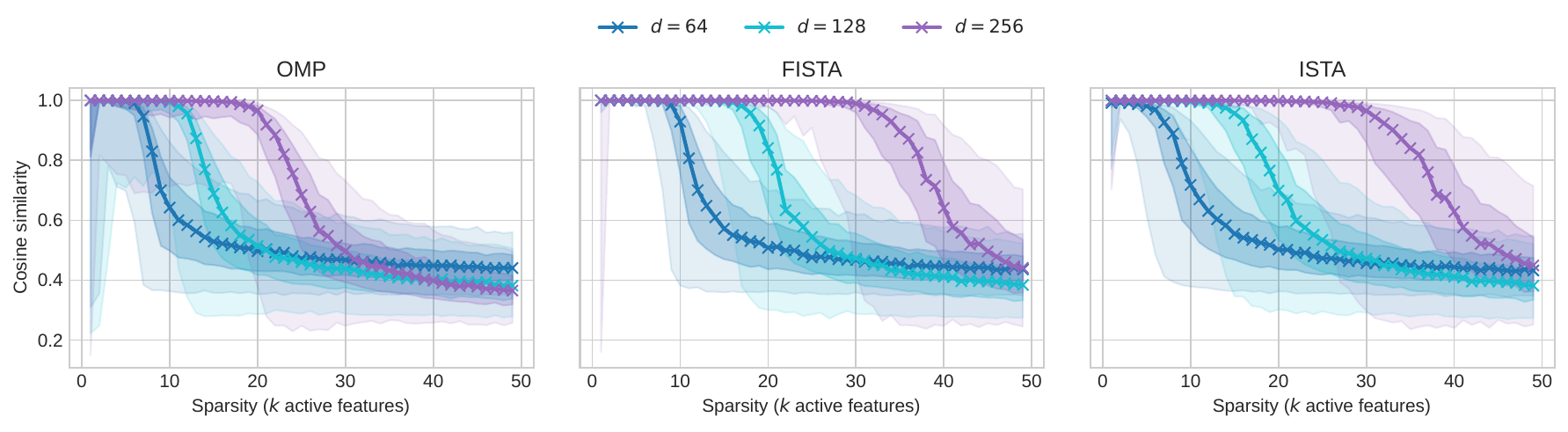}
        \captionof{figure}{Dictionary recovery using OMP, FISTA and ISTA,
            where the last two optimize a LASSO objective.
            Figures present the median, 5th, 25th, 75th, and 95th percentiles
            of the matched cosine similarities between corresponding ground-truth and learned dictionary columns in a noiseless setting.
            Feature directions and their coefficients are drawn from a normal distribution.
            Details in Section~\ref{sec:dictionary_learning}.}
        \vspace{-10pt}
        \label{fig:dictionary_learning}
    \end{figure}

    \newpage

    \section{Feature recovery using SAEs}\label{sec:methodology_sae}
    To measure SAE performance across varying sparsities ($k \in [1, 50)$) and noise levels ($\{0, 0.1\}$)
    for a dictionary of size $m=1024$ embedded in $d=128$ dimensions, we train models for $10,000$ steps using batches of $1024$ samples.
    Samples are generated as $Y = D_{\text{true}}X + z$ (general vector space),
    optionally followed by RMSNorm,
    where the number of active features is drawn uniformly from $[1, k]$ during training
    and fixed to $k$ during evaluation,
    and $z$ is Gaussian noise scaled by the signal variance.
    Active coefficients are sampled from $|x| \sim |\mathcal{N}(0,1)| + 0.1$.

    We evaluate an iterative TopK SAE (unfolded with learned step sizes) and MP-SAE (utilizing sequential matching pursuit residual updates)
    across $n \in \{0, 1, 2, 4, 8, 16, 32\}$ iterations, where $n=0$ is equivalent to a standard single-step TopK SAE\@.
    Models are optimized using Adam (learning rate $10^{-3}$, cosine annealing) to minimize mean squared error
    alongside an auxiliary residual-based reconstruction loss for dead latent revival.
    The decoder dictionary columns are $\ell_2$-normalized after every parameter update.
    Performance is evaluated across $2048$ test samples using the F1-score for active support recovery.

    \subsection{Optimal encoder baseline}
    \label{subsec:optimal_encoder_baseline}
    We compute a coefficient vector $c_S$ that maximizes the separation margin between active and inactive features for a single-step matched-filter readout.
    For a true active signal $x_S$ with support $S$ and active dictionary columns $A_S$,
    we optimize $c_S$ such that the minimum active readout score exceeds the maximum inactive score by a target margin $\gamma$.
    Given the readout scores $s = A^\top A_S c_S$, the objective minimizes the margin violation:

    $$\min_{c_S} \max(0, \gamma - (\min_{i \in S} s_i - \max_{j \notin S} s_j))$$

    We solve this objective iteratively using the Adam optimizer with a learning rate of 0.01 for 200 iterations, setting the target margin $\gamma=0.1$.

    Simply put, when the true coefficient $x_j$ is $1$, but is measured $1 - 1 = 0$ because of interference,
    we add $1$ to the transmitted signal so it measures $2 - 1 = 1$, which is the ground truth.
    However, increasing $c_j$ might change the measurement for another active or inactive feature.
    This optimizes the coefficients for all active and inactive features simultaneously.

    \subsection{SAE in general vector space}\label{subsec:appendix_sae_general_vector_space}

    \begin{figure}[h!]
        \centering
        \includegraphics[width=\linewidth]{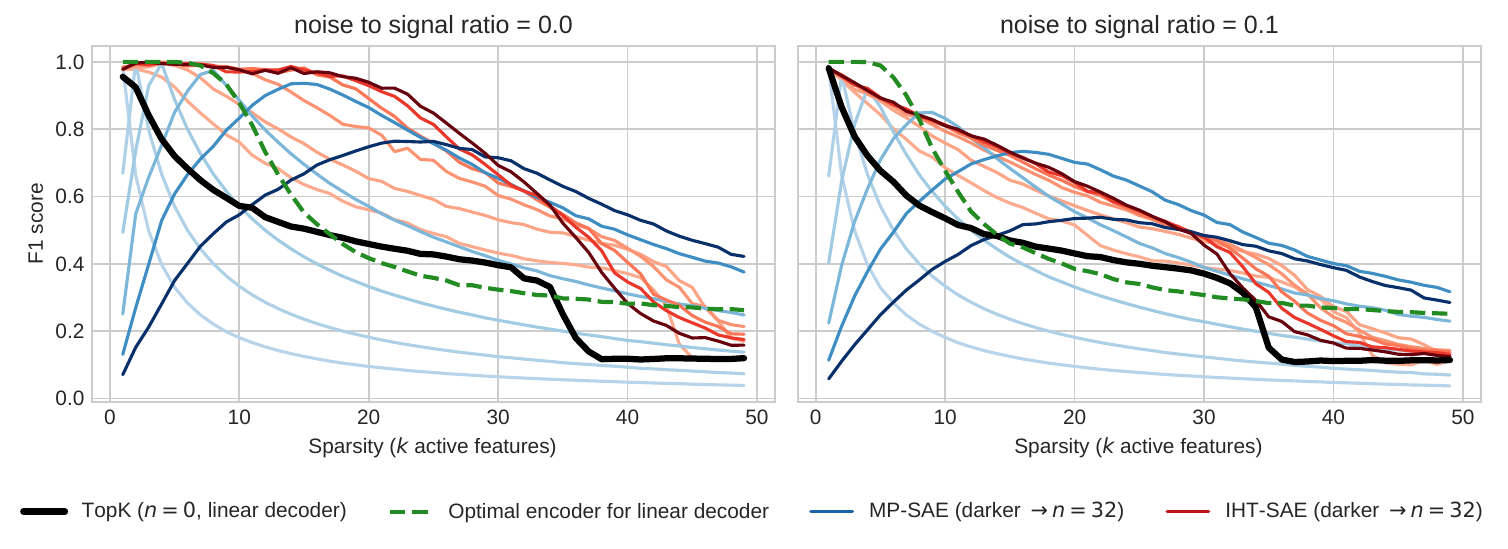}
        \captionof{figure}{Support recovery using TopK, IHT-SAE and MP-SAE in general vector space ($m=1024$, $d=128$).
        Curves show F1 scores versus active sparsity $k$ at noise-to-signal power ratios of $0$ and $0.1$.
        Darker shades indicate more iterations ($n \in \{1,2,4,8,16,32\}$); black denotes single-step TopK ($n=0$).}
        \vspace{-10pt}
    \end{figure}

    \newpage

    \subsection{IHT-SAE algorithm}
    \label{sec:appendix_IHT_sae}

    \begin{figure}[h!]
        \centering
        \begin{minipage}{0.95\linewidth}
            \hrule
            \medskip
            \textbf{TopK-SAE and IHT-SAE}
            \medskip
            \hrule
            \medskip

            \textbf{Input:} $x$; encoder $W_e$; decoder $D$; biases
            $b_{\mathrm{pre}}, b_{\mathrm{lat}}$; sparsity budget $k$;
            IHT step sizes $\{\eta_t\}_{t=1}^{T}$.

            \begin{tabbing}
                \quad\=\quad\=\kill
                \textbf{function} $\operatorname{TopKReLU}(v,k)$\\
                \> $S \gets$ extract indices of the $k$ largest entries in $v$\\
                \> $h \gets$ initialize zero vector of the same shape as $v$\\
                \> $h_S \gets \operatorname{ReLU}(v_S)$\\
                \> \textbf{return} $h$\\
                \textbf{end function}\\[6pt]

                \textbf{TopK-SAE}\\
                $y \gets x - b_{\mathrm{pre}}$\\
                $h_0 \gets \operatorname{TopKReLU}(W_e y + b_{\mathrm{lat}}, k)$\\
                $\hat{x}_{\mathrm{TopK}} \gets Dh_0 + b_{\mathrm{pre}}$\\[6pt]

                \textbf{IHT-SAE}\\
                $h \gets h_0$\\
                \textbf{for} $n = 1,\ldots,N$ \textbf{do}\\
                \> $r \gets y - Dh$\\
                \> $h \gets \operatorname{TopKReLU}(h + \eta_n D^\top r, k)$\\
                \textbf{end for}\\
                $\hat{x}_{\mathrm{IHT}} \gets Dh + b_{\mathrm{pre}}$
            \end{tabbing}
            \hrule
        \end{minipage}
        \caption{TopK-SAE and IHT-SAE forward passes.
        IHT-SAE refines a TopK initialization through residual-based updates;
            $N=0$ recovers TopK-SAE.
            TopK selects the largest signed values before applying ReLU,
            retaining at most $k$ active coefficients.}
        \label{fig:sae_algorithms}
    \end{figure}

    \newpage

    \section{Gaussian approximation on other distributions}
    \label{sec:appendix_gaussian_approximation_on_other_distributions}

    \begin{figure}[H]
        \centering
        \begin{subfigure}{\textwidth}
            \centering
            \includegraphics[width=\textwidth]{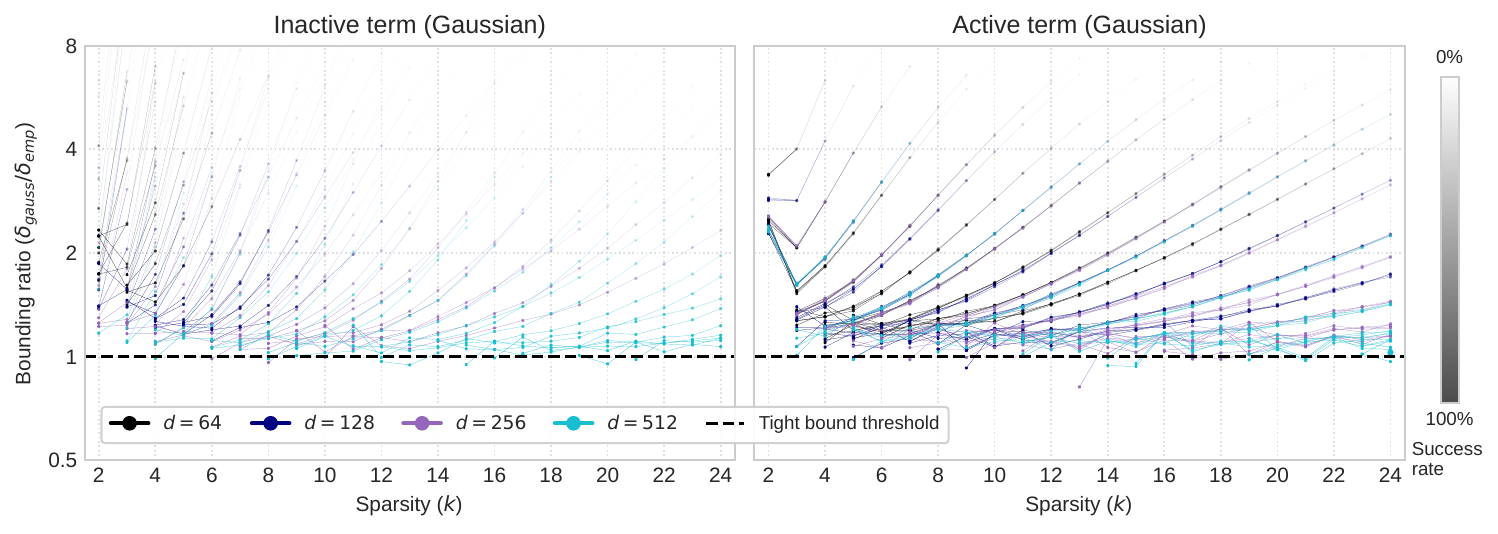}
            \caption{Gaussian distribution}
            \label{fig:bound_gaussian}
        \end{subfigure}
        \vspace{1em}
        \begin{subfigure}{\textwidth}
            \centering
            \includegraphics[width=\textwidth]{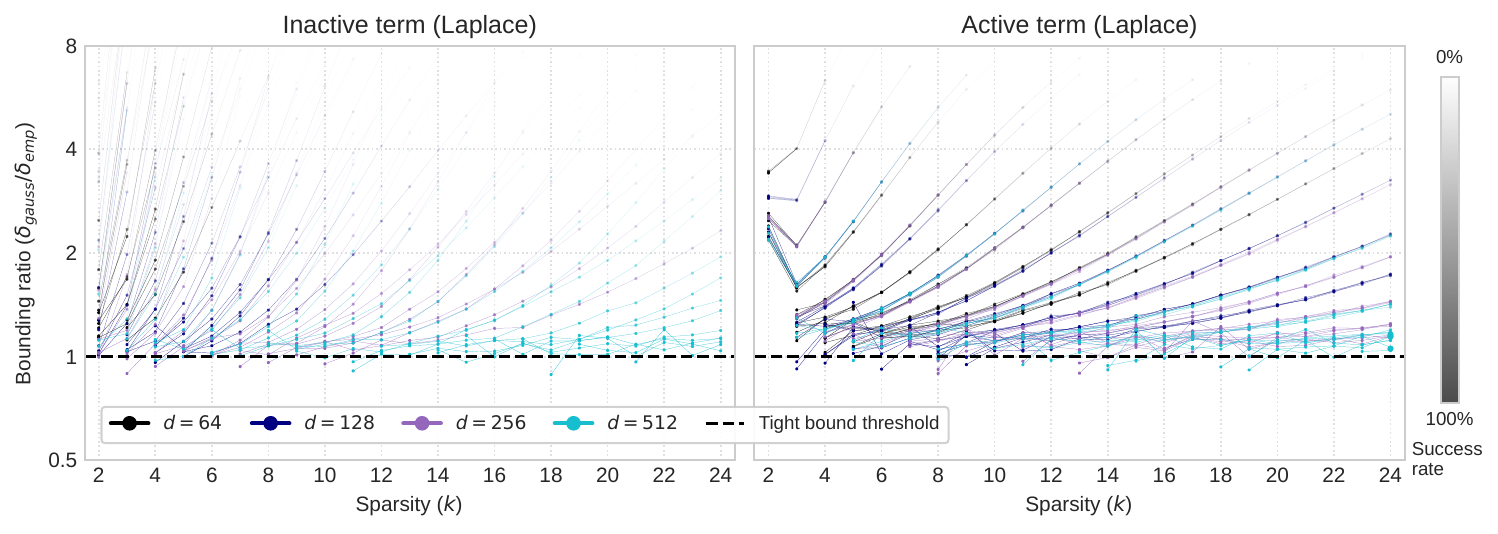}
            \caption{Laplace distribution}
            \label{fig:bound_laplace}
        \end{subfigure}

        \caption{Tightness of Gaussian approximation for high-probability decoding on other distributions.
            $40,000$ trials were run for every combination of sparsity $k \in [2, 24]$,
            dictionary size $m \in \{512, 1024, 2048, 4096\}$ and dimension $d \in \{64, 128, 256, 512\}$,
            evaluated against various error thresholds $\varepsilon \in \{0.1, 0.2, \dots, 0.9\}$.}
        \label{fig:gaussian_laplace_bound}
    \end{figure}

\end{document}

%% file: math_commands.tex
\usepackage{amsmath,amsfonts,bm}

\def\eqref#1{equation~\ref{#1}}

\def\1{\bm{1}}

\DeclareMathAlphabet{\mathsfit}{\encodingdefault}{\sfdefault}{m}{sl}
\SetMathAlphabet{\mathsfit}{bold}{\encodingdefault}{\sfdefault}{bx}{n}



%% file: linear_sufficient_average_case.tex
\section{Sufficient high-probability bound for linear decoding}\label{sec:appendix_average_case_proof}

Grounded in the linear representation hypothesis,
the dictionary $A = [a_1, \dots, a_m] \in \mathbb{R}^{d \times m}$ encodes semantic features as normalized latent directions ($\|a_i\|_2 = 1$).
The observed state is a sparse linear superposition of these vectors $y = Ax + z$, where $z$ is additive observation noise.
We analyze the linear decoder:
\begin{equation}
    \widehat{x} = A^\top y
\end{equation}

Let $d, m, k \in \mathbb{N}_{>0}$ with $1 \le k \le m$,
where $d$ is the ambient dimension (number of observations), $m$ the total feature count, and $k$ the maximum sparsity.
Fix an arbitrary deterministic index set $S \subseteq [m]$ with sparsity $1 \le | S | \le k$.
We make the following assumptions throughout:
\begin{itemize}[noitemsep, topsep=0pt, parsep=2pt, partopsep=0px, leftmargin=*]
\item[-] Dictionary columns $a_1, \dots, a_m$ are mean-zero, and sampled independently and uniformly from the unit sphere $S^{d-1}$.
\item[-] The active signal coefficients $(x_j)_{j \in S}$ are drawn from any joint distribution $P_x \in \mathcal{P}_S$,
where $\mathcal{P}_S$ denotes the broad class of all valid bounded coefficient distributions ($| x_j | \le 1$) that are independent of $A$.
We set $x_j=0$ for $j \notin S$.
\item[-] Observation noise $z = (z_1, \dots, z_d)$ is drawn from a distribution $P_z \in \mathcal{Z}_\sigma$,
where $\mathcal{Z}_\sigma$ denotes the class of product noise distributions independent of $(A, x)$
with independent, mean-zero coordinates satisfying the subgaussian norm bound $\|z_n\|_{\psi_2} \le \sigma$, where $\sigma>0$.
\end{itemize}

Our analysis of this decoding draws largely upon the high-dimensional probability techniques detailed in~\cite{vershynin2026highdimensional},
from which the theorems utilized below are sourced.
We fixed a support $S$ to avoid the uniform-support combinatorial penalty,
with inactive coordinates set to zero ($x_j=0$ for $j \notin S$).
For any target coordinate $i \in [m]$, the number of interfering active features is denoted by $s_i := | S \setminus \{i\} |$.
Because $S$ is fixed,
$s_i$ is a fixed quantity bounded by $s_i = | S | - 1$ when $i \in S$,
and $s_i = | S |$ when $i \notin S$.
In either case, $s_i \le | S | \le k$.

\subsection{Error decomposition}\label{subsec:error_decomposition}

Let $\varepsilon > 0$ denote the maximum acceptable global decoding error.
To analyze this error for a target coordinate $i$,
we expand the observation model $y = \sum_{j \in S} x_j a_j + z$ and subtract the true signal $x_i$ from the decoder estimate $\widehat{x}_i = \langle a_i, y \rangle$:
\begin{equation}
    \begin{aligned}
        \widehat{x}_i - x_i &= \langle a_i, y \rangle - x_i \\
        &= \left\langle a_i, \sum_{j \in S} x_j a_j + z \right\rangle - x_i \\
        &= x_i \langle a_i, a_i \rangle + \sum_{j \in S \setminus \{i\}} x_j \langle a_i, a_j \rangle + \langle a_i, z \rangle - x_i \\
        &= x_i(\langle a_i, a_i \rangle - 1) + \underbrace{\sum_{j \in S \setminus \{i\}} x_j \langle a_i, a_j \rangle}_{I_i} + \underbrace{\langle a_i, z \rangle}_{N_i}
    \end{aligned}
\end{equation}

Because the dictionary vectors are normalized ($\|a_i\|_2=1$), the self-term vanishes ($x_i(1 - 1) = 0$),
isolating the decoding error to the cross-feature interference $I_i$ and the projected noise $N_i$:
\begin{equation}
    \widehat{x}_i - x_i = I_i + N_i
\end{equation}

We define $\mathcal{I} := \{i : s_i \ge 1\}$ as the subset of coordinates that face cross-talk (worst-case having a cardinality of $m$);
for $i \notin \mathcal{I}$, the interference is zero ($s_i = 0$ and $I_i = 0$).
To separate the two contributions to the total error, we partition the global tolerance $\varepsilon$.
We allocate a fraction $\alpha \in (0, 1)$ to the noise, and $(1-\alpha)$ to the interference.
By the triangle inequality, if the total error exceeds $\varepsilon$, at least one component must necessarily exceed its allocated budget.
This bounds the total failure event within the union of individual failures:
\begin{equation}
    \{|I_i + N_i| \ge \varepsilon\} \subseteq \{|I_i| \ge (1-\alpha)\varepsilon\} \cup \{|N_i| \ge \alpha\varepsilon\}
\end{equation}
Applying the union bound to these sets yields the decoupled probability bound:
\begin{equation}
    \mathbb{P}(|I_i + N_i| \ge \varepsilon) \le \mathbb{P}(|I_i| \ge (1-\alpha)\varepsilon) + \mathbb{P}(|N_i| \ge \alpha\varepsilon)
\end{equation}

\subsection{Interference concentration}\label{subsec:interference_concentration}

We evaluate the interference concentration in the joint probability space of the dictionary and the signal.
For any fixed target coordinate $i \in \mathcal{I}$ and any distribution $P_x \in \mathcal{P}_S$,
with $S$ fixed, we condition on the signal $x$ and the target vector $a_i$ realizations.
Recall $I_i = \sum_{j\in S\setminus\{i\}} x_j U_j$, where $U_j = \langle a_i, a_j \rangle$.
Since $a_j \sim \text{Unif}(S^{d-1})$ and $a_i$ is a fixed unit vector,
evaluating all moments allows us to bound the exponential moment.

To do so, we count the ways to partition $2r$ factors into disjoint pairs.
Through the sphere version of Wick's theorem~\citep{vignat2008wick},
this count gives the even moments of $U_j$,
which we then bound and substitute into the exponential series to bound its exponential moment.
Let $I_{2r}=\{1,\ldots,2r\}$ and let $\Pi_{2r}$ be the set of partitions of $I_{2r}$ into $r$ disjoint pairs.
For each integer $r\ge1$,
$U_j^{2r}=\prod_{k=1}^{2r}\langle a_i,a_j\rangle=\langle a_i,a_j\rangle\cdot\langle a_i,a_j\rangle\cdots\langle a_i,a_j\rangle$,
with $2r$ identical factors.
Because $a_i$ is fixed and the expectation is over $a_j$,
Wick's theorem pairs the $a_i$ from one factor with the $a_i$ from another,
resolving every pair to $\langle a_i, a_i \rangle$.
For a pairing $\sigma\in\Pi_{2r}$, let $I_{2r}/\sigma$ contain one representative from each pair, giving $r$ indices in total.
Then:
\begin{equation}
    \mathbb{E}_A[U_j^{2r}\mid a_i]
    =\frac{\Gamma(d/2)}{2^r\Gamma(r+d/2)}
    \underbrace{\sum_{\sigma\in\Pi_{2r}}}_{\lvert\Pi_{2r}\rvert\text{ terms}}
    \underbrace{\prod_{k\in I_{2r}/\sigma}\langle a_i,a_i\rangle}_{r\text{ factors, all }1}
    =\frac{1\cdot3\cdots(2r-1)}{d(d+2)\cdots(d+2r-2)}
    \le \frac{(2r)!}{2^r r!\,d^r}
    \label{eq:wick}
\end{equation}

Here $\Gamma$ is the Gamma function, satisfying $\Gamma(z+1)=z\Gamma(z)$,
and $\lvert\Pi_{2r}\rvert$ denotes the number of pairings.
Each pairing contributes $1$ because $\langle a_i,a_i\rangle=\|a_i\|_2^2=1$.
The final equality uses
$\lvert\Pi_{2r}\rvert=1\cdot3\cdots(2r-1)$ and the Gamma
recurrence to obtain the denominator $d(d+2)\cdots(d+2r-2)$.
The inequality follows by bounding this denominator below
by $d^r$ and using
$1\cdot3\cdots(2r-1)=(2r)!/(2^r r!)$.

To control the probability of large interference via the exponential moment method,
we first bound $\mathbb{E}_A[e^{\lambda U_j}\mid a_i]$.
Since $|U_j|\le1$, we can expand the exponential and average its series term by term.
The odd moments vanish by symmetry, and the even-moment bound above yields for every $\lambda\in\mathbb{R}$
(we will choose the exact value later):
\begin{align}
    \mathbb{E}_A[e^{\lambda U_j}\mid a_i]
    &=
    1+\frac{\lambda^2}{2!}\mathbb{E}_A[U_j^2\mid a_i]
    +\frac{\lambda^4}{4!}\mathbb{E}_A[U_j^4\mid a_i]
    +\frac{\lambda^6}{6!}\mathbb{E}_A[U_j^6\mid a_i]
    +\cdots \\
    &=1+\sum_{r=1}^{\infty}\frac{\lambda^{2r}}{(2r)!}\mathbb{E}_A[U_j^{2r}\mid a_i] \\
    &\le1+\sum_{r=1}^{\infty}\frac{\lambda^{2r}}{(2r)!}\frac{(2r)!}{2^r r!\,d^r} \\
    &=1+\sum_{r=1}^{\infty}\frac{1}{r!}\left(\frac{\lambda^2}{2d}\right)^r
    =\exp\left(\frac{\lambda^2}{2d}\right)
    \label{eq:sphere_exponential_moment}
\end{align}

Conditional on $x,a_i$, the interference $I_i$ is symmetric about zero,
so its two tails have equal probability.
For $t>0$ and $\lambda>0$, we apply Markov's inequality (Proposition~1.6.2) to $e^{\lambda I_i}$ and multiply by two.
Conditional independence factors the expectation into a product,
and the exponential-moment bound with parameter $\lambda x_j$ bounds each factor.
Thus:
\begin{equation}
    \begin{aligned}
        \mathbb{P}_A(|I_i|\ge t\mid x,a_i)
        &=2\mathbb{P}_A(I_i\ge t\mid x,a_i)
        =2\mathbb{P}_A(e^{\lambda I_i}\ge e^{\lambda t}\mid x,a_i) \\
        &\le2\frac{\mathbb{E}_A[e^{\lambda I_i}\mid x,a_i]}{e^{\lambda t}}
        =2e^{-\lambda t}\prod_{j\in S\setminus\{i\}}\mathbb{E}_A[e^{\lambda x_j U_j}\mid x,a_i]
    \end{aligned}
    \label{eq:markov_product_bound}
\end{equation}

Substituting Equation~\ref{eq:sphere_exponential_moment},
with parameter $\lambda x_j$, into Equation~\ref{eq:markov_product_bound} yields:
\begin{equation}
    \begin{aligned}
        \mathbb{P}_A(|I_i|\ge t\mid x,a_i)
        &\le 2e^{-\lambda t}
        \prod_{j\in S\setminus\{i\}}
        \exp\left(\frac{(\lambda x_j)^2}{2d}\right) \\
        &=2e^{-\lambda t}
        \exp\left(
                \sum_{j\in S\setminus\{i\}}
                \frac{\lambda^2x_j^2}{2d}
        \right) \\
        &=2\exp\left(
                   -\lambda t+\frac{\lambda^2}{2d}
                   \sum_{j\in S\setminus\{i\}}x_j^2
        \right).
    \end{aligned}
    \label{eq:hoeffding_inequality}
\end{equation}

Using $|x_j|\le1$ and $\sum_{j\in S\setminus\{i\}}x_j^2\le s_i$, we obtain the variance proxy:
\begin{equation}
    \frac{1}{d}\sum_{j\in S\setminus\{i\}}x_j^2\le\frac{s_i}{d}
    \label{eq:variance_proxy}
\end{equation}

For $s_i>0$, choosing $\lambda=dt/s_i$ minimizes the quadratic exponent $-\lambda t+\lambda^2s_i/(2d)$.

Thus, substituting Equation~\ref{eq:variance_proxy} into Equation~\ref{eq:hoeffding_inequality}
and choosing $\lambda=dt/s_i$, we obtain:
\begin{equation}
    \mathbb{P}_A\left(| I_i|\ge t\mid x,a_i\right)
    \le 2\exp\left(-\frac{dt^2}{s_i} + \frac{dt^2}{2s_i}\right)
    = 2\exp\left(-\frac{dt^2}{2s_i}\right)
\end{equation}

We did this instead of applying the subgaussian Hoeffding inequality (Theorem 2.7.3) to avoid an unspecified absolute constant by proving it is $0.5$.
Furthermore, Theorem 2.2.1 assumes Rademacher variables, so we couldn't use it directly, but the steps in this proof follow the same exponential moment method.

Because the right hand side of this bound is deterministic and independent of the specific realizations of $x$ and $a_i$,
taking the expectation over $x \sim P_x$ and $a_i \sim \text{Unif}(S^{d-1})$ preserves the bound in the joint $(A,x)$ probability space:
\begin{equation}
    \mathbb{P}_{A,x}\left(|I_i|\ge t\right) \le 2\exp\left(-\frac{dt^2}{2s_i}\right)
    \label{eq:hoeffding_envelope}
\end{equation}

Intuitively, if the failure probability satisfies this bound for every individual realization of the signal and target vector,
the expected failure probability across all random signals and vectors preserves this exact same bound.
As such, to bound the worst-case cross-talk across all valid coordinates $i \in \mathcal{I}$,
we seek the probability that the maximum interference exceeds our allocated threshold $t = (1-\alpha)\varepsilon$.
Applying the union bound over $\mathcal{I}$ transforms the maximum into a sum of individual tail probabilities:
\begin{equation}
    \mathbb{P}_{A,x}\left(\max_{i\in\mathcal I}|I_i|\ge(1-\alpha)\varepsilon\right)
    \le \sum_{i \in \mathcal{I}} \mathbb{P}_{A,x}\left(| I_i|\ge(1-\alpha)\varepsilon\right)
    \label{eq:max_bounded_by_sum}
\end{equation}

Where $\max_{i\in\varnothing}|I_i|=0$.
We evaluate this sum by partitioning $\mathcal{I}$ into inactive coordinates ($i \notin S$, where $s_i = |S|$) and active coordinates ($i \in S$,
where $s_i = |S|-1$). This splits the sum into two components,
yielding the joint interference constraint with failure probability at most $\delta_{\text{int}}$:
\begin{equation}
    \begin{aligned}
        &\mathbb{P}_{A,x}\left(\max_{i\in\mathcal I}|I_i|\ge(1-\alpha)\varepsilon\right)\\
        &\quad \le \underbrace{2(m-|S|)\exp\left(-\frac{d(1-\alpha)^2\varepsilon^2}{2|S|}\right)}_{\text{inactive}}
        +\underbrace{\mathbb{I}_{|S| \ge 2} \cdot 2|S|\exp\left(-\frac{d(1-\alpha)^2\varepsilon^2}{2(|S|-1)}\right)}_{\text{active}} \\
        &\quad \le2m\exp\left(-\frac{d(1-\alpha)^2\varepsilon^2}{2|S|}\right)\le\delta_{\text{int}}
    \end{aligned}
\end{equation}

The \textit{inactive} term controls the interference budget on inactive coordinates,
where the interference is generated by all $s_i = |S|$ active features,
whose erroneous inclusion would produce false positives.
The \textit{active} term controls the interference budget on active coordinates,
where the interference is generated by the remaining $s_i = |S|-1$ active features,
whose erroneous exclusion would produce false negatives.
Here, the term vanishes when $|S| = 1$.
The union bound includes both terms, so the resulting condition controls the
interference failures relevant to both error types.
Inverting the rightmost inequality for $d$ yields the Lemma~\ref{lemma:interference_constraint}.

\begin{figure}[H]
    \centering
    \includegraphics[width=\textwidth]{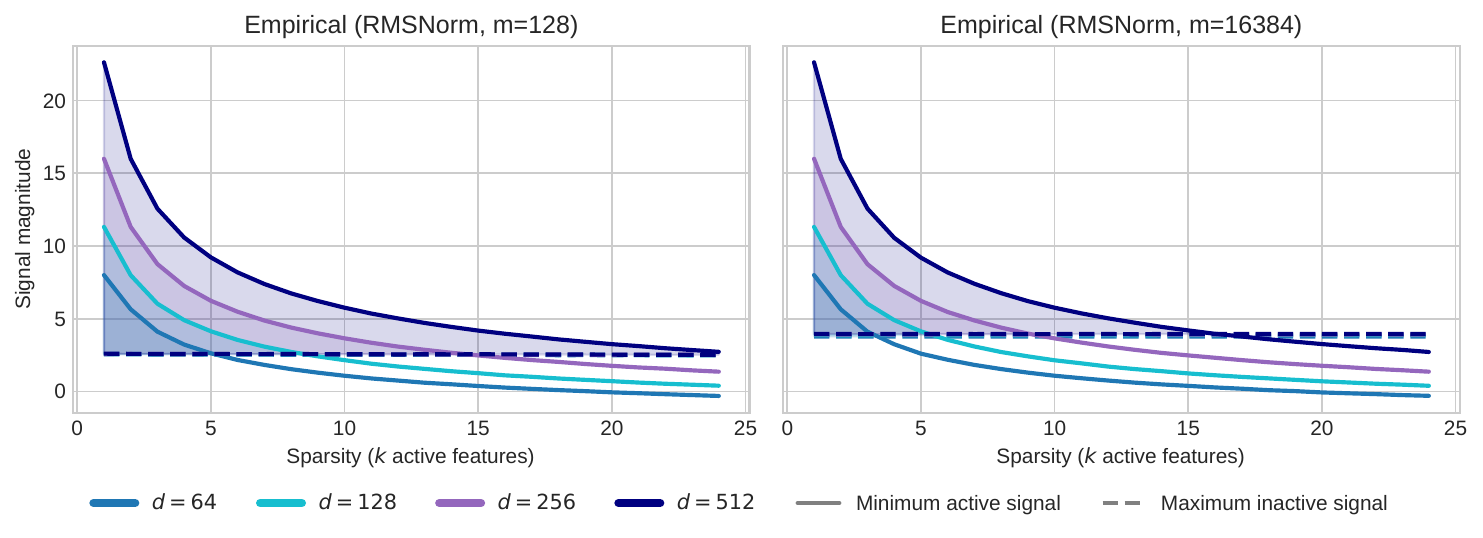}
    \vspace{-20pt}
    \caption{Support recovery in an RMSNorm vector space for different dictionary sizes in a noiseless setting.
        Consistent with the theoretical high-probability bound $d = O_\epsilon(k \log m)$,
        the empirical recovery gap visualizes that linear accessibility supports exponentially many features.
        Success rates are evaluated across varying ambient dimensions $d$ and sparsities $k$.}
    \label{fig:success_m_values}
\end{figure}

\subsection{Gaussian approximation of $\delta_{\text{int}}$}\label{subsec:theoretical_delta}

Recall that interference $I_i = \sum_{j\in S\setminus\{i\}} x_j U_j$,
where $U_j = \langle a_i, a_j \rangle$, is a linear combination of 1D projections.
Since noise is absent, we simplify $(1-\alpha)\varepsilon = \varepsilon$.
By the projective central limit theorem (Theorem 3.3.9),
1D marginals of vectors uniformly distributed on the unit sphere converge to a normal distribution.
Because $|x_j| \le 1$, summing the independent projection variances of $1/d$ across the $s_i$ active interfering features
results in a worst-case total interference variance of $\sigma^2 = s_i / d$.
Standardizing $I_i / \sigma \approx \mathcal{N}(0, 1)$ and defining the normalized threshold $t = \varepsilon / \sigma = \varepsilon \sqrt{d / s_i}$,
we apply the Gaussian tail bound (Proposition 2.1.2) to approximate the two-sided failure probability:

\begin{equation}
    \begin{aligned}
        \mathbb{P}(|I_i| \ge \varepsilon) = \mathbb{P}(|I_i / \sigma| \ge t) &\approx 2 \left( \frac{1}{t \sqrt{2\pi}} \right) e^{-t^2 / 2} \\
        &= 2 \left( \frac{\sqrt{s_i}}{\varepsilon \sqrt{2\pi d}} \right) \exp\left( -\frac{d \varepsilon^2}{2s_i} \right)
    \end{aligned}
\end{equation}

For comparison, we define a tail-matching factor by matching
the Gaussian approximation to the form of the Hoeffding envelope (Equation~\ref{eq:hoeffding_envelope}).
By allocating the entire error budget to the interference term,
the threshold for the Hoeffding envelope becomes $t = \varepsilon$.
Subsequently isolating a tail-matching factor $c_{\text{tail}}$ yields:
\begin{equation}
    \begin{aligned}
        2\exp\left(-c_{\text{tail}} \frac{d \varepsilon^2}{s_i}\right) &= 2 \left( \frac{\sqrt{s_i}}{\varepsilon \sqrt{2\pi d}} \right) \exp\left( -\frac{d \varepsilon^2}{2s_i} \right) \\
        -c_{\text{tail}} \frac{d \varepsilon^2}{s_i} &= \ln \left( \frac{\sqrt{s_i}}{\varepsilon \sqrt{2\pi d}} \right) - \frac{d \varepsilon^2}{2s_i} \\
        c_{\text{tail}} &= 0.5 - \frac{s_i}{d \varepsilon^2} \ln \left( \frac{\sqrt{s_i}}{\varepsilon \sqrt{2 \pi d}} \right)
    \end{aligned}
    \label{eq:tail_matching_factor}
\end{equation}

To understand how an absolute constant can bound the tail-matching factor from below,
let's substitute $\varepsilon \sqrt{d / s_i} = t$ back, yielding:
\begin{equation}
    c_{\text{tail}} = 0.5 - \frac{1}{t^{2}} \ln \left( \frac{1}{t \sqrt{2 \pi}} \right)
\end{equation}

\begin{figure}[H]
    \centering
    \includegraphics[width=0.9\textwidth]{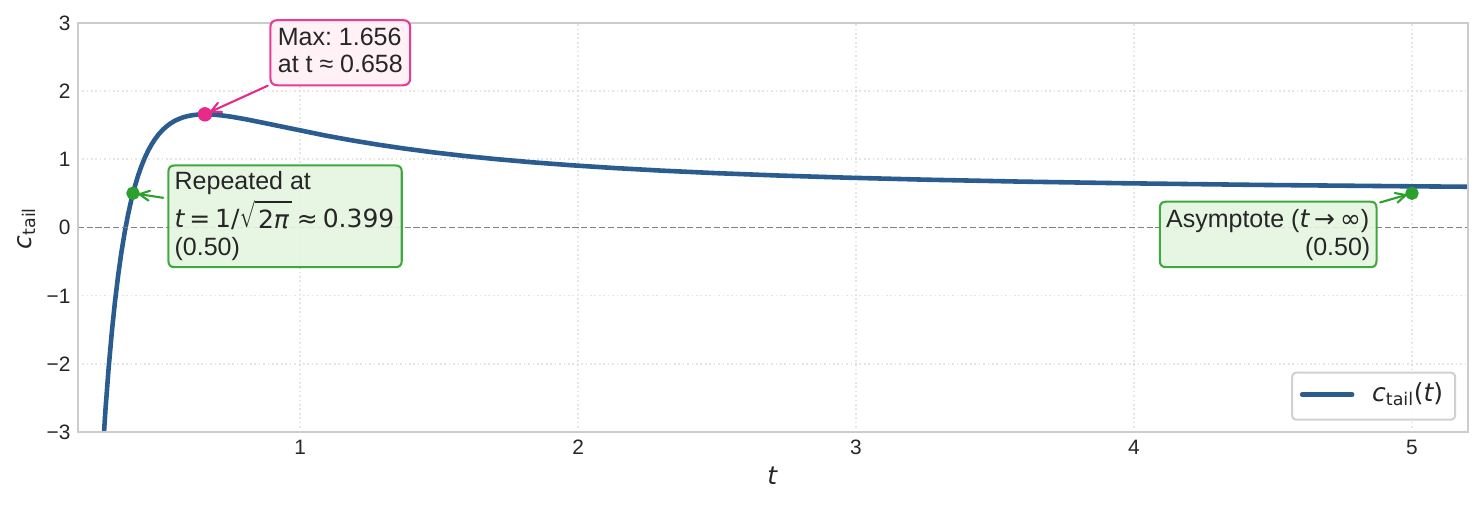}
    \caption{$c_{\text{tail}}(t)$ is bounded from below by $0.5$ for $t = \varepsilon \sqrt{d / s_i} \ge 1/\sqrt{2\pi}$.}
    \label{fig:c_tail}
\end{figure}

To compute the Gaussian approximation for $\delta_{\text{int}}$,
we plug $c_{\text{tail}}$ back into the Hoeffding envelope,
and apply the union bound $M \in \{(m-|S|), |S|\}$ over the exponent:
\begin{equation}
    \delta_{\text{gauss}}
    = 2M \exp\left(- \max\left(0.5, 0.5 - \frac{s_i}{d \varepsilon^2} \ln \left( \frac{\sqrt{s_i}}{\varepsilon \sqrt{2 \pi d}} \right)\right) \frac{d \varepsilon^2}{s_i}\right)
    \label{eq:gaussian_approximation}
\end{equation}

This approximation captures the scaling dynamics of the system parameters.
The floor of $0.5$ within the exponent ensures that the numerical estimate does not exceed the baseline Hoeffding bound.
The $c_{\text{tail}}$ term is then utilized to occasionally get a better numerical estimate,
yielding more accurate numerical results in regimes where the Gaussian tail decays faster than the worst-case Hoeffding envelope.
In configurations where the number of features is small relative to the dimension size, the exponential decay yields a high probability of successful decoding;
and instead when dimension size is relatively smaller, the probability decreases accordingly.

\subsection{Noise concentration}\label{subsec:noise_concentration}

Conditional on $A$, because the projected noise $N_i = \langle a_i, z \rangle = \sum_{n=1}^d a_{i,n} z_n$ is a sum of independent, mean-zero subgaussian variables,
we can apply the subgaussian Hoeffding inequality (Theorem 2.7.3), where $c > 0$ is a constant:
\begin{equation}
    \mathbb{P}_z\left(| N_i|\ge t \mid A\right) \le 2\exp\left(-\frac{c t^2}{\sum_{n=1}^d \|a_{i,n} z_n\|_{\psi_2}^2}\right)
    \label{eq:noise_hoeffding}
\end{equation}

By the definition of our noise class $\mathcal{Z}_\sigma$, the subgaussian norm of each coordinate is $\|z_n\|_{\psi_2} \le \sigma$.
When factoring out $a_{i,n}$ and using the fact that $a_i$ is a unit vector ($\|a_i\|_2=1$), we get:
\begin{equation}
    \sum_{n=1}^d \|a_{i,n} z_n\|_{\psi_2}^2 \le \sigma^2 \sum_{n=1}^d a_{i,n}^2 = \sigma^2
    \label{eq:noise_subgaussian_sum}
\end{equation}

Substituting Equation~\ref{eq:noise_subgaussian_sum} into Equation~\ref{eq:noise_hoeffding} with our allocated error threshold $t = \alpha\varepsilon$, we obtain:
\begin{equation}
    \mathbb{P}_z\left(| N_i|\ge \alpha\varepsilon \mid A\right) \le 2\exp\left(-\frac{c\alpha^2\varepsilon^2}{\sigma^2}\right)
    \label{eq:hoeffding_noise_envelope}
\end{equation}

Applying a union bound over all $m$ coordinates,
analogous to Equation~\ref{eq:max_bounded_by_sum}, yields the conditional noise constraint:
\begin{equation}
    \mathbb{P}_z\left(\max_{i\in[m]}| N_i|\ge\alpha\varepsilon\,\middle|\,A\right)
    \le\sum_{i=1}^m\mathbb{P}_z\left(| N_i|\ge\alpha\varepsilon\,\middle|\,A\right)
    \le2m\exp\left(-\frac{c\alpha^2\varepsilon^2}{\sigma^2}\right)\le\delta_{\text{noise}}
\end{equation}

Since this conditional bound holds uniformly for every realization of $A$,
it also holds under the joint distribution of $(A,z)$,
with failure probability at most $\delta_{\text{noise}}$.
Inverting the rightmost inequality for $\sigma$ yields the Lemma~\ref{lemma:noise_constraint}.

%% file: references.bib
@book{vershynin2026highdimensional,
    author = {Vershynin, Roman},
    title = {High-Dimensional Probability: An Introduction with Applications in Data Science},
    edition = {2},
    publisher = {Cambridge University Press},
    address = {Cambridge},
    series = {Cambridge Series in Statistical and Probabilistic Mathematics},
    year = {2026}
}

@inproceedings{ba2011lower,
    title = {Lower Bounds for Sparse Recovery},
    author = {Khanh Do Ba and Piotr Indyk and Eric Price and David P. Woodruff},
    year = {2010},
    booktitle = {SODA},
    url = {https://arxiv.org/abs/1106.0365},
}

@inproceedings{garg2026featureslanguagemodelstore,
    title = {How Many Features Can a Language Model Store Under the Linear Representation Hypothesis?},
    author = {Nikhil Garg and Jon Kleinberg and Kenny Peng},
    year = {2026},
    booktitle = {COLT},
    url = {https://arxiv.org/abs/2602.11246},
}

@article{candes2006stable,
    author = {Candès, Emmanuel J. and Romberg, Justin K. and Tao, Terence},
    title = {Stable signal recovery from incomplete and inaccurate measurements},
    journal = {Communications on Pure and Applied Mathematics},
    volume = {59},
    number = {8},
    pages = {1207-1223},
    url = {https://onlinelibrary.wiley.com/doi/abs/10.1002/cpa.20124},
    year = {2006}
}

@article{candes2008ripimplications,
    title = {The restricted isometry property and its implications for compressed sensing},
    journal = {Comptes Rendus Mathematique},
    volume = {346},
    number = {9},
    pages = {589-592},
    year = {2008},
    issn = {1631-073X},
    url = {https://www.sciencedirect.com/science/article/pii/S1631073X08000964},
    author = {Emmanuel J. Candès},
}

@inproceedings{candes2005decoding,
    title = {Decoding by Linear Programming},
    author = {Emmanuel Candes and Terence Tao},
    booktitle = {IEEE Transactions on Information Theory},
    year = {2005},
    url = {https://arxiv.org/abs/math/0502327},
}

@inproceedings{pacela2026stop,
    title = {Stop Probing, Start Coding: Why Linear Probes and Sparse Autoencoders Fail at Compositional Generalisation},
    author = {Vit{\'o}ria Barin-Pacela and Shruti Joshi and Isabela Camacho and Simon Lacoste-Julien and David Klindt},
    booktitle = {2nd Workshop on Compositional Learning: Safety, Interpretability, and Agents},
    year = {2026},
    url = {https://openreview.net/forum?id=3NN3ehMKxW}
}

@book{rubinstein2016montecarlo,
    author = {Rubinstein, Reuven Y. and Kroese, Dirk P.},
    date = {2016-11-21},
    title = {Simulation and the Monte Carlo Method},
    url = {https://app.dimensions.ai/details/publication/pub.1106805670},
    year = {2016}
}

@inproceedings{liu2025on,
    title = {On the Learn-to-Optimize Capabilities of Transformers in In-Context Sparse Recovery},
    author = {Renpu Liu and Ruida Zhou and Cong Shen and Jing Yang},
    booktitle = {ICLR},
    year = {2025},
    url = {https://openreview.net/forum?id=NHhjczmJjo}
}

@article{zhao2023beyond,
    title = {Beyond Single Concept Vector: Modeling Concept Subspace in LLMs with Gaussian Distribution},
    author = {Zhao, Haiyan and Zhao, Heng and Shen, Bo and Payani, Ali and Yang, Fan and Du, Mengnan},
    journal = {ICLR},
    year = {2025},
    url = {https://arxiv.org/abs/2410.00153},
}

@article{sidorov2020,
    title = {Linear and Fisher Separability of Random Points in the d-Dimensional Spherical Layer and Inside the d-Dimensional Cube},
    journal = {IJCNN},
    author = {Sidorov, Sergey and Zolotykh, Nikolai},
    year = {2020},
}

@inproceedings{stevinson2025adversarial,
    title = {Adversarial Attacks Leverage Interference Between Features in Superposition},
    author = {Edward Stevinson and Lucas Prieto and Melih Barsbey and Tolga Birdal},
    booktitle = {Mechanistic Interpretability Workshop at NeurIPS 2025},
    year = {2025},
    url = {https://openreview.net/forum?id=LqI52GG2Ss}
}

@inproceedings{prieto2026from,
    title = {From Data Statistics to Feature Geometry: How Correlations Shape Superposition},
    author = {Lucas Prieto and Edward Stevinson and Melih Barsbey and Tolga Birdal and Pedro A. M. Mediano},
    booktitle = {The Fourteenth International Conference on Learning Representations},
    year = {2026},
    url = {https://openreview.net/forum?id=7akSRQS5Xh}
}

@article{radford2021learning,
    author = {Radford, Alec and Kim, Jong Wook and Hallacy, Chris and Ramesh, Aditya and Goh, Gabriel and Agarwal, Sandhini and Sastry, Girish and Askell, Amanda and Mishkin, Pamela and Clark, Jack and Krueger, Gretchen and Sutskever, Ilya},
    title = {Learning Transferable Visual Models From Natural Language Supervision},
    year = {2021},
    journal = {ICML}
}

@inproceedings{park2024lrh,
    author = {Park, Kiho and Choe, Yo Joong and Veitch, Victor},
    title = {The linear representation hypothesis and the geometry of large language models},
    year = {2024},
    booktitle = {ICML},
    url = {https://arxiv.org/abs/2311.03658},
}

@inproceedings{park2025the,
    title = {The Geometry of Categorical and Hierarchical Concepts in Large Language Models},
    author = {Kiho Park and Yo Joong Choe and Yibo Jiang and Victor Veitch},
    booktitle = {ICLR},
    year = {2025},
    url = {https://openreview.net/forum?id=bVTM2QKYuA}
}

@inproceedings{ethayarajh2019contextual,
    title = "How Contextual are Contextualized Word Representations? {C}omparing the Geometry of {BERT}, {ELM}o, and {GPT}-2 Embeddings",
    author = "Ethayarajh, Kawin",
    booktitle = "EMNLP-IJCNLP",
    year = "2019",
    url = "https://aclanthology.org/D19-1006/",
}

@inproceedings{cai2021isotropy,
    title = {Isotropy in the Contextual Embedding Space: Clusters and Manifolds},
    author = {Xingyu Cai and Jiaji Huang and Yuchen Bian and Kenneth Church},
    booktitle = {International Conference on Learning Representations},
    year = {2021},
    url = {https://openreview.net/forum?id=xYGNO86OWDH}
}

@misc{klindt2026does,
    title = {When Does LeJEPA Learn a World Model?},
    author = {David Klindt and Yann LeCun and Randall Balestriero},
    year = {2026},
    eprint = {2605.26379},
    archivePrefix = {arXiv},
    primaryClass = {stat.ML},
    url = {https://arxiv.org/abs/2605.26379},
}

@misc{vompa2026scaling,
    title = {The Scaling Properties of Implicit Deductive Reasoning in Transformers},
    author = {Enrico Vompa and Tanel Tammet},
    year = {2026},
    archivePrefix = {arXiv},
    url = {https://arxiv.org/abs/2605.04330},
}

@article{vompa2026beyond,
    title = {Beyond the Linear Separability Ceiling: Aligning Representations in {VLM}s},
    author = {Enrico Vompa and Tanel Tammet and Mohit Vaishnav},
    journal = {Transactions on Machine Learning Research},
    year = {2026},
    url = {https://openreview.net/forum?id=3uX4p80bN0}
}

@article{thomas2021hypercomputing,
    title = {A Theoretical Perspective on Hyperdimensional Computing},
    volume = {72},
    url = {http://dx.doi.org/10.1613/jair.1.12664},
    journal = {Journal of Artificial Intelligence Research},
    author = {Thomas, Anthony and Dasgupta, Sanjoy and Rosing, Tajana},
    year = {2021},
    month = Oct
}

@article{elhage2022superposition,
    title = {Toy Models of Superposition},
    author = {Elhage, Nelson and Hume, Tristan and Olsson, Catherine and Schiefer, Nicholas and Henighan, Tom and Kravec, Shauna and Hatfield-Dodds, Zac and Lasenby, Robert and Drain, Dawn and Chen, Carol and Grosse, Roger and McCandlish, Sam and Kaplan, Jared and Amodei, Dario and Wattenberg, Martin and Olah, Christopher},
    year = {2022},
    journal = {Transformer Circuits Thread},
    url = {https://transformer-circuits.pub/2022/toy_model/index.html}
}

@article{lindeberg1922,
    author = {Lindeberg, J. W.},
    journal = {Mathematische Zeitschrift},
    pages = {211-225},
    title = {Eine neue Herleitung des Exponentialgesetzes in der Wahrscheinlichkeitsrechnung},
    url = {http://eudml.org/doc/167717},
    volume = {15},
    year = {1922},
}

@inbook{zhang2019rmsnorm,
    author = {Zhang, Biao and Sennrich, Rico},
    title = {Root mean square layer normalization},
    year = {2019},
    booktitle = {NeurIPS},
    url = {https://arxiv.org/abs/1910.07467},
}

@article{vignat2008wick,
    title = {An extension of Wick’s theorem},
    journal = {Statistics and Probability Letters},
    year = {2008},
    url = {https://www.sciencedirect.com/science/article/pii/S0167715208001405},
    author = {C. Vignat and S. Bhatnagar},
}

@article{beck2009fista,
    author = {Beck, Amir and Teboulle, Marc},
    title = {A Fast Iterative Shrinkage-Thresholding Algorithm for Linear Inverse Problems},
    journal = {SIAM Journal on Imaging Sciences},
    volume = {2},
    number = {1},
    year = {2009},
    URL = {https://doi.org/10.1137/080716542},
}

@article{daubechies2004ista,
    author = {Daubechies, I. and Defrise, M. and De Mol, C.},
    title = {An iterative thresholding algorithm for linear inverse problems with a sparsity constraint},
    journal = {Communications on Pure and Applied Mathematics},
    url = {https://onlinelibrary.wiley.com/doi/abs/10.1002/cpa.20042},
    year = {2004}
}

@article{robert1996lasso,
    author = {Tibshirani, Robert},
    title = {Regression Shrinkage and Selection Via the Lasso},
    journal = {Journal of the Royal Statistical Society: Series B (Methodological)},
    year = {1996},
    month = {01},
    issn = {0035-9246},
    url = {https://doi.org/10.1111/j.2517-6161.1996.tb02080.x},
}

@article{blumensath2009hard,
    title = {Iterative hard thresholding for compressed sensing},
    journal = {Applied and Computational Harmonic Analysis},
    year = {2009},
    issn = {1063-5203},
    url = {https://doi.org/10.1016/j.acha.2009.04.002},
    author = {Thomas Blumensath and Mike E. Davies},
}

@inproceedings{pati1993omp,
    author = {Pati, Y.C. and Rezaiifar, R. and Krishnaprasad, P.S.},
    booktitle = {Proceedings of 27th Asilomar Conference on Signals, Systems and Computers},
    title = {Orthogonal matching pursuit: recursive function approximation with applications to wavelet decomposition},
    year = {1993},
    url = {https://ieeexplore.ieee.org/document/342465}
}

@inproceedings{engan1999mod,
    author = {Engan, K. and Aase, S.O. and Hakon Husoy, J.},
    booktitle = {1999 IEEE International Conference on Acoustics, Speech, and Signal Processing. Proceedings. ICASSP99 (Cat. No.99CH36258)},
    title = {Method of optimal directions for frame design},
    year = {1999},
    url = {https://ieeexplore.ieee.org/document/760624} }

@inproceedings{gregor2010lista,
    author = {Gregor, Karol and LeCun, Yann},
    title = {Learning fast approximations of sparse coding},
    year = {2010},
    booktitle = {Proceedings of the 27th International Conference on International Conference on Machine Learning},
    url = {https://dl.acm.org/doi/10.5555/3104322.3104374}
}

@inproceedings{costa2025from,
    title = {From Flat to Hierarchical: Extracting Sparse Representations with Matching Pursuit},
    author = {Val{\'e}rie Costa and Thomas Fel and Ekdeep Singh Lubana and Bahareh Tolooshams and Demba E. Ba},
    booktitle = {NeurIPS},
    year = {2025},
    url = {https://openreview.net/forum?id=Ll5miDx8KB}
}

@misc{gao2024topk,
    title = {Scaling and evaluating sparse autoencoders},
    author = {Leo Gao and Tom Dupré la Tour and Henk Tillman and Gabriel Goh and Rajan Troll and Alec Radford and Ilya Sutskever and Jan Leike and Jeffrey Wu},
    year = {2024},
    url = {https://arxiv.org/abs/2406.04093},
}

@article{engels2025not,
    author = {Engels, Joshua and Michaud, Eric J and Liao, Isaac and Gurnee, Wes and Tegmark, Max},
    title = {Not All Language Model Features Are One-Dimensionally Linear},
    year = {2025},
    journal = {ICLR}
}

@misc{kantamneni2025helix,
    title = {Language Models Use Trigonometry to Do Addition},
    author = {Subhash Kantamneni and Max Tegmark},
    year = {2025},
    booktitle = {ICLR Workshop},
    url = {https://arxiv.org/abs/2502.00873},
}

@inproceedings{huben2024sparse,
    title = {Sparse Autoencoders Find Highly Interpretable Features in Language Models},
    author = {Robert Huben and Hoagy Cunningham and Logan Riggs Smith and Aidan Ewart and Lee Sharkey},
    booktitle = {ICLR},
    year = {2024},
    url = {https://openreview.net/forum?id=F76bwRSLeK}
}

@article{wen2020coefficient,
    title = {Signal-Dependent Performance Analysis of Orthogonal Matching Pursuit for Exact Sparse Recovery},
    url = {http://dx.doi.org/10.1109/TSP.2020.3016571},
    journal = {IEEE Transactions on Signal Processing},
    author = {Wen, Jinming and Zhang, Rui and Yu, Wei},
    year = {2020},
}

@misc{bai2023qwentechnicalreport,
    title = {Qwen Technical Report},
    author = {Jinze Bai and Shuai Bai and Yunfei Chu and Zeyu Cui and Kai Dang and Xiaodong Deng and Yang Fan and Wenbin Ge and Yu Han and Fei Huang and Binyuan Hui and Luo Ji and Mei Li and Junyang Lin and Runji Lin and Dayiheng Liu and Gao Liu and Chengqiang Lu and Keming Lu and Jianxin Ma and Rui Men and Xingzhang Ren and Xuancheng Ren and Chuanqi Tan and Sinan Tan and Jianhong Tu and Peng Wang and Shijie Wang and Wei Wang and Shengguang Wu and Benfeng Xu and Jin Xu and An Yang and Hao Yang and Jian Yang and Shusheng Yang and Yang Yao and Bowen Yu and Hongyi Yuan and Zheng Yuan and Jianwei Zhang and Xingxuan Zhang and Yichang Zhang and Zhenru Zhang and Chang Zhou and Jingren Zhou and Xiaohuan Zhou and Tianhang Zhu},
    year = {2023},
    eprint = {2309.16609},
    archivePrefix = {arXiv},
    primaryClass = {cs.CL},
    url = {https://arxiv.org/abs/2309.16609},
}

@InProceedings{parkhi12a,
    author = "Omkar M. Parkhi and Andrea Vedaldi and Andrew Zisserman and C. V. Jawahar",
    title = "Cats and Dogs",
    booktitle = "IEEE Conference on Computer Vision and Pattern Recognition",
    year = "2012",
}

@misc{eldan2023tiny,
    title={TinyStories: How Small Can Language Models Be and Still Speak Coherent English?},
    author={Ronen Eldan and Yuanzhi Li},
    year={2023},
    eprint={2305.07759},
    archivePrefix={arXiv},
    primaryClass={cs.CL},
    url={https://arxiv.org/abs/2305.07759},
}

@article{bricken2023monosemanticity,
    title = {Towards Monosemanticity: Decomposing Language Models With Dictionary Learning},
    author = {Bricken, Trenton and Templeton, Adly and Batson, Joshua and Chen, Brian and Jermyn, Adam and Conerly, Tom and Turner, Nick and Anil, Cem and Denison, Carson and Askell, Amanda and Lasenby, Robert and Wu, Yifan and Kravec, Shauna and Schiefer, Nicholas and Maxwell, Tim and Joseph, Nicholas and Hatfield-Dodds, Zac and Tamkin, Alex and Nguyen, Karina and McLean, Brayden and Burke, Josiah E and Hume, Tristan and Carter, Shan and Henighan, Tom and Olah, Christopher},
    year = {2023},
    journal = {Transformer Circuits Thread},
    note = {https://transformer-circuits.pub/2023/monosemantic-features/index.html}
}
